\documentclass[preprint, 12pt]{elsarticle}

\usepackage{amssymb}
\usepackage{amsmath}
\usepackage{hyperref}
\usepackage{graphicx}
\usepackage{caption}
\usepackage{subfigure}
\usepackage{array}
\usepackage{multirow}

\journal{}
\biboptions{sort&compress}
\begin{document}
\begin{frontmatter}


\title{Three-Stream Temporal-Shift Attention Network Based on Self-Knowledge Distillation for Micro-Expression Recognition}


\author[a]{Guanghao Zhu} 
\author[a]{Lin Liu} 
\author[a]{Yuhao Hu} 
\author[a]{Haixin Sun} 
\author[a]{Fang Liu} 
\author[a]{Xiaohui Du} 
\author[a]{Ruqian Hao} 
\author[a]{Juanxiu Liu} 
\author[b]{Yong Liu} 
\author[a]{Jing Zhang\corref{cor1}} 
\cortext[cor1]{Corresponding author}
\ead{zhangjing@uestc.edu.cn}

\address[a]{MOEMIL Laboratory, School of Optoelectronic Science and Engineering, University of Electronic Science and Technology of China, Chengdu, 611731, China}
\address[b]{School of Optoelectronic Science and Engineering, University of Electronic Science and Technology of China, Chengdu, 611731, China}

\begin{abstract}
Micro-expressions are subtle facial movements that occur spontaneously when people try to conceal real emotions. Micro-expression recognition is crucial in many fields, including criminal analysis and psychotherapy. However, micro-expression recognition is challenging since micro-expressions have low intensity and public datasets are small in size. To this end, a three-stream temporal-shift attention network based on self-knowledge distillation is proposed in this paper. Firstly, to address the low intensity of muscle movements, we utilize learning-based motion magnification modules to enhance the intensity of muscle movements. Secondly, we employ efficient channel attention modules in the local-spatial stream to make the network focus on facial regions that are highly relevant to micro-expressions. In addition, temporal shift modules are used in the dynamic-temporal stream, which enables temporal modeling with no additional parameters by mixing motion information from two different temporal domains. Furthermore, we introduce self-knowledge distillation into the micro-expression recognition task by introducing auxiliary classifiers and using the deepest section of the network for supervision, encouraging all blocks to fully explore the features of the training set. Finally, extensive experiments are conducted on five publicly available micro-expression datasets. The experimental results demonstrate that our network outperforms other existing methods and achieves new state-of-the-art performance. Our code is available at \href{https://github.com/GuanghaoZhu663/SKD-TSTSAN}{{https://github.com/GuanghaoZhu663/SKD-TSTSAN}}.
\end{abstract}




\end{frontmatter}



\section{Introduction}
\label{sec1}

Facial expression (FE) is a common means of nonverbal communication and is crucial for understanding human mental states and intentions \citep{li2022deep}. Facial expressions can generally be categorized into two classes: macro-expressions and micro-expressions (MEs). The major variations between them are intensity and duration. Macro-expressions typically have a duration of 0.5-4 seconds and involve muscle movements covering a large area of the face, which can easily be recognized by humans \citep{matsumoto2011evidence}. In contrast, MEs reflect hidden emotions that come out spontaneously when people strive to hide their true emotions. It usually lasts only 1/25 to 1/3 of a second with muscle movements in a minor region of the face \citep{ekman2009lie, li2025could}, which is difficult to recognize, and only an expert with extensive training can distinguish MEs. MEs are significant in many fields, such as crime analysis \citep{polikovsky2009facial}, emotional communication, and clinical diagnosis. Therefore, researchers are striving to utilize computer vision techniques for automatic micro-expression recognition (MER).

Early MER methods are based on manual feature extraction, such as Local Binary Pattern histograms from Three Orthogonal Planes (LBP-TOP) \citep{pfister2011recognising} and Main Directional Mean Optical-flow (MDMO) \citep{liu2015main}. These methods largely rely on manually designed extractors and complicated parameter adjustments. The extracted features are mainly low-level features \citep{li2022deep}, therefore it is difficult to recognize the subtle movements in MEs with high accuracy. With the proposal of a series of ME datasets \citep{yan2014casme, davison2016samm, li2013spontaneous, ben2021video, li2022cas, li20224dme} and the development of deep learning, researchers began to use CNN to extract spatial and temporal features of MEs. For example, Song et al.~\cite{song2019recognizing} proposed a three-stream convolutional neural network (TSCNN) to learn the spatiotemporal information of MEs. Xia and Wang~\cite{xia2021micro} trained two dual-stream baseline models called MiNet and MaNet using the ME dataset and macro-expression dataset. By introducing two auxiliary tasks between the two models, MiNet can effectively extract both the dynamic patterns of the muscle motion and the static texture of the face. Recent studies have explored more structured spatiotemporal modeling for MER, e.g., coupling graph-based relational reasoning with Transformer-style feature learning, and enhancing subtle motion representation with multi-scale attention mechanisms \citep{li2024micro, he2025micro_nest}. Although these methods can extract ME features and achieve promising results, there are still two major challenges in MER: the low intensity of MEs and overfitting on small ME datasets \citep{ben2021video}. Since MEs have low intensity, networks need to have strong representation ability, which generally means more convolutional layers and computational costs. However, complex models usually overfit on small-scale ME datasets. Therefore, in order to mine subtle ME motions from small ME datasets, more effective blocks need be introduced.

To address the above challenges, we propose a three-stream temporal-shift attention network based on self-knowledge distillation called SKD-TSTSAN in this paper. Firstly, we use the TSCNN \citep{song2019recognizing} as our baseline, which consists of three streams, namely static-spatial stream (S-stream), local-spatial stream (L-stream), and dynamic-temporal stream (T-stream). The inputs of the three streams are the whole facial image, the concatenation of $4\times 4$ localized facial patches (16 patches per face), and optical flow images, respectively. The three recognition streams are finally fused into a fully connected layer for MER tasks. 

To tackle the low intensity of MEs, we propose a stream-specific integration of motion magnification \citep{oh2018learning}: a dedicated module is placed at the head of each stream, where the S-stream module amplifies whole-face apex frames, the L-stream module amplifies local facial patches, and the T-stream module amplifies optical flow fields. This end-to-end integration—without a reconstruction decoder—enhances subtle muscle motions across all input modalities, making the features of each stream more discriminative. Since MEs typically appear in localized facial regions, such as the mouth and eyebrows, the input to the L-stream often contains redundant information. To address this, we incorporate efficient channel attention (ECA) modules \citep{wang2020eca} into the L-stream, enabling the network to focus on meaningful local regions while ignoring redundant information.

Compared with 2D CNNs, 3D CNNs can mine the spatiotemporal information of videos more effectively \citep{li2022deep}. However, 3D CNNs generally have a large number of parameters and are easily overfitted on small-scale ME datasets \citep{li2022deep}. To effectively model temporal dynamics without the computational overhead of 3D CNNs, we introduce temporal shift modules (TSMs) \citep{lin2019tsm} in T-stream. These modules enable efficient information exchange along the temporal dimension without introducing additional parameters. To further enhance the performance of our lightweight network, the macro-expression dataset CK+ \citep{lucey2010extended} is employed for pre-training. Moreover, we employ self-knowledge distillation (SKD) \citep{zhang2019your}. Traditional knowledge distillation requires a complex teacher model to guide a compact student model  \citep{hinton2015distilling}, but designing an effective over-parameterized teacher model is challenging on small-scale micro-expression datasets. In SKD, we introduce additional convolutional and fully connected layers to form multiple auxiliary classifiers (ACs). By training ACs using both labels and the model's final output, we ensure that all layers of the network comprehensively learn the training set's features.

In summary, the main contributions of this paper are as follows:
\begin{itemize}
    \item We propose a stream-specific integration strategy for motion magnification, where dedicated modules are independently applied to each stream's input modality within a unified end-to-end framework, amplifying subtle muscle motions across all feature streams without a reconstruction decoder.
    \item ECA modules are introduced in the local-spatial stream to focus on meaningful facial regions while ignoring redundant information.
    \item TSMs are integrated into the dynamic-temporal stream, enabling efficient temporal modeling without additional parameters.
    \item We explore the effectiveness of self-knowledge distillation on the MER tasks, introducing auxiliary classifiers in the network to enable comprehensive feature learning from the training set.
    \item Extensive experiments on five ME datasets demonstrate that SKD-TSTSAN achieves state-of-the-art (SOTA) performance. Ablation studies further validate the effectiveness of each module.
\end{itemize}

\section{Related Work}
\label{sec2}

In this section, we initially introduce current micro-expression recognition methods, including hand-crafted methods and deep learning methods. Then, since our proposed SKD-TSTSAN uses temporal shift modules (TSMs) \citep{lin2019tsm} for temporal modeling in MER and incorporates the self-knowledge distillation (SKD) method \citep{zhang2019your}, we also briefly introduce temporal modeling and knowledge distillation.

\subsection{Micro-expression recognition}
\label{subsec2-1}

\subsubsection{Hand-crafted methods}
\label{subsubsec2-1-1}

Hand-crafted methods commonly extract manually designed appearance-based ME features and provide them to classic classifiers for emotion recognition, including the K nearest neighbor (KNN) classifier \citep{chaudhry2009histograms, zhang2017micro} and support vector machine (SVM) \citep{wang2015micro, xu2017microexpression}. Pfister et al.~\cite{pfister2011recognising} employed a temporal interpolation model (TIM) to obtain a sufficient amount of frames, and then utilized the LBP-TOP as the texture descriptor for classification. Since then, many LBP-TOP-based enhanced MER methods have been proposed. For example, Wang et al.~\cite{wang2015lbp} proposed Local Binary Patterns with Six Intersection Points (LBP-SIP) descriptor, which replaces the three orthogonal planes containing redundant points with intersecting lines of them. Only six different neighboring points around the center point are used to obtain the spatiotemporal LBP patterns, which improves the computational efficiency. In addition to LBP-TOP features, optical flow features, which reflect the motion patterns of moving objects, are also significant in ME research \citep{ben2021video}. Liu et al.~\cite{liu2015main} proposed an MDMO feature that only considers mean optical flow in the main direction, thus having fewer feature dimensions. In conclusion, although these approaches offer strong interpretability, they depend on specialized knowledge and extensive parameter adjustments \citep{ben2021video}. Besides, the extracted features are relatively shallow, which is not conducive to recognizing subtle movements in ME.

\subsubsection{Deep-learning methods}
\label{subsubsec2-1-2}

Deep learning-based MER methods typically combine feature extraction and classification, achieving SOTA predictive performance in recent years. Since optical flow can reflect object motion by detecting changes in pixel intensity between two frames, it has been widely applied in MER tasks. For example, Wang et al.~\cite{wang2024htnet} presented a hierarchical transformer network (HTNet), which first calculates the optical flow between the onset frame and the apex frame, and divides the face into four different facial regions. The transformer layers capture fine-grained features within the local regions, and aggregation blocks facilitate interaction between different regions. VTSFNet \citep{wang2025micro} integrates visual features with fine-grained Action Unit textual semantics. It employs a local region attention mechanism to capture dependencies in optical flow images. Furthermore, the model utilizes a cross-modal attention mechanism to align these visual cues with BERT-encoded AU descriptions.

Due to the low intensity of ME, some researchers have utilized motion magnification techniques to enhance muscle movements in ME sequences \citep{lei2020novel, zhang2022motion, wei2022novel}, such as Eulerian Motion Magnification (EMM). Traditional motion magnification techniques relying on hand-crafted filters often result in noise and excessive blurring. Therefore, Oh et al.~\cite{oh2018learning} proposed a learning-based motion magnification network that directly learns filters from data, achieving high-quality magnification results. Subsequently, learning-based motion magnification algorithm has been widely used in MER tasks. For example, Wei et al.~\cite{wei2022novel} employed a set of different amplification factors (AFs) to amplify ME movements and used an attention mechanism to assign varying weights to these AFs, adaptively focusing on the appropriate amplification level.

The ME datasets are relatively small in scale, while macro-expression datasets are large and share some common features, such as similar Action Units (AUs) when expressing emotions \citep{ben2021video}. Therefore, using transfer learning to transfer knowledge from macro-expressions to MEs can typically enhance the performance of MER methods. For example, Wang et al.~\cite{wang2018micro} employed a two-step transfer learning approach. Firstly, they pre-trained the CNN using a large-scale dataset of general expressions. Then, they fine-tuned the pre-trained network by treating each frame of the ME video clips as a sample. This approach leverages the high relevance between the source and target domains, ensuring improved performance. 

Although deep learning-based MER methods have shown promising results, extracting discriminative representations from limited ME samples remains challenging. More recently, Zhang et al.~\cite{zhang2025dynamic} proposed a deformation-based approach guided by dynamic stereotype theory, which models the oriented deformation of facial muscle movements for improved ME representation. Therefore, more efficient modules need to be developed.

\subsection{Temporal Modeling}
\label{subsec2-2}

Using 3D CNNs is a direct approach for temporal modeling. Li et al.~\cite{li2019micro} proposed a 3D flow-based CNN composed of three data stream subnetworks. 3D convolutional kernels were employed to extract spatiotemporal feature information. However, using deep 3D CNNs significantly increases computational costs. Thus, Xie et al.~\cite{xie2020assisted} adopted a lightweight 3D CNN backbone called STPNet to extract spatiotemporal features, where the 3D convolutional filter is decomposed into 2D spatial convolution and 1D temporal convolution. Wu and Guo~\cite{wu2021tsnn} proposed a network for MER that combines 2D CNNs and 3D CNNs. This network uses 3D convolutional kernels with varying settings to extract features in different temporal domains and then utilizes 2D CNN blocks to extract spatial features.

Another approach for modeling temporal relations is using 2D CNN + post-hoc fusion. Khor et al.~\cite{khor2018enriched} proposed an Enriched Long-term Recurrent Convolutional Network (ELRCN), where the ME features are fed into a Long Short-Term Memory (LSTM) network \citep{gers2000learning} to learn the temporal sequence information of MEs. However, 3D CNN-based methods have high computational requirements and tend to overfit on limited ME datasets, while 2D CNN + post-hoc fusion methods cannot capture low-level information that is lost in the feature extraction process. Therefore, we introduce temporal shift modules (TSMs) \citep{lin2019tsm} into the dynamic-temporal stream of our three-stream network to capture the temporal relationships with no additional parameters.

\subsection{Knowledge Distillation}
\label{subsec2-3}

Knowledge distillation is an effective approach for compressing models. The key is to make a compact student model approximate an over-parameterized teacher model, thereby achieving significant performance improvements \citep{hinton2015distilling}. Using a shallow student model instead of a teacher model enables compression and acceleration. Sun et al.~\cite{sun2020dynamic} employed a pre-trained network for AU detection as the teacher network, extracting knowledge from AUs and transferring it to a shallow student network. However, in traditional two-stage distillation, designing an appropriate teacher model is challenging. Moreover, excellent student models that outperform teacher models are rare due to inefficient knowledge transfer \citep{zhang2019your}. 

To address the shortcomings of traditional distillation, we employ the self-distillation framework proposed by \cite{zhang2019your}, which adds multiple auxiliary classifiers (AC) to the model. The deepest part and the shallow part with classifiers are respectively treated as the teacher model and the student model, achieving single-stage self-knowledge distillation (SKD). Experimental results show that using SKD improves network performance while maintaining low complexity.


\section{Method}
\label{sec3}

\begin{figure}[t]
\centering
\includegraphics[width=\textwidth]{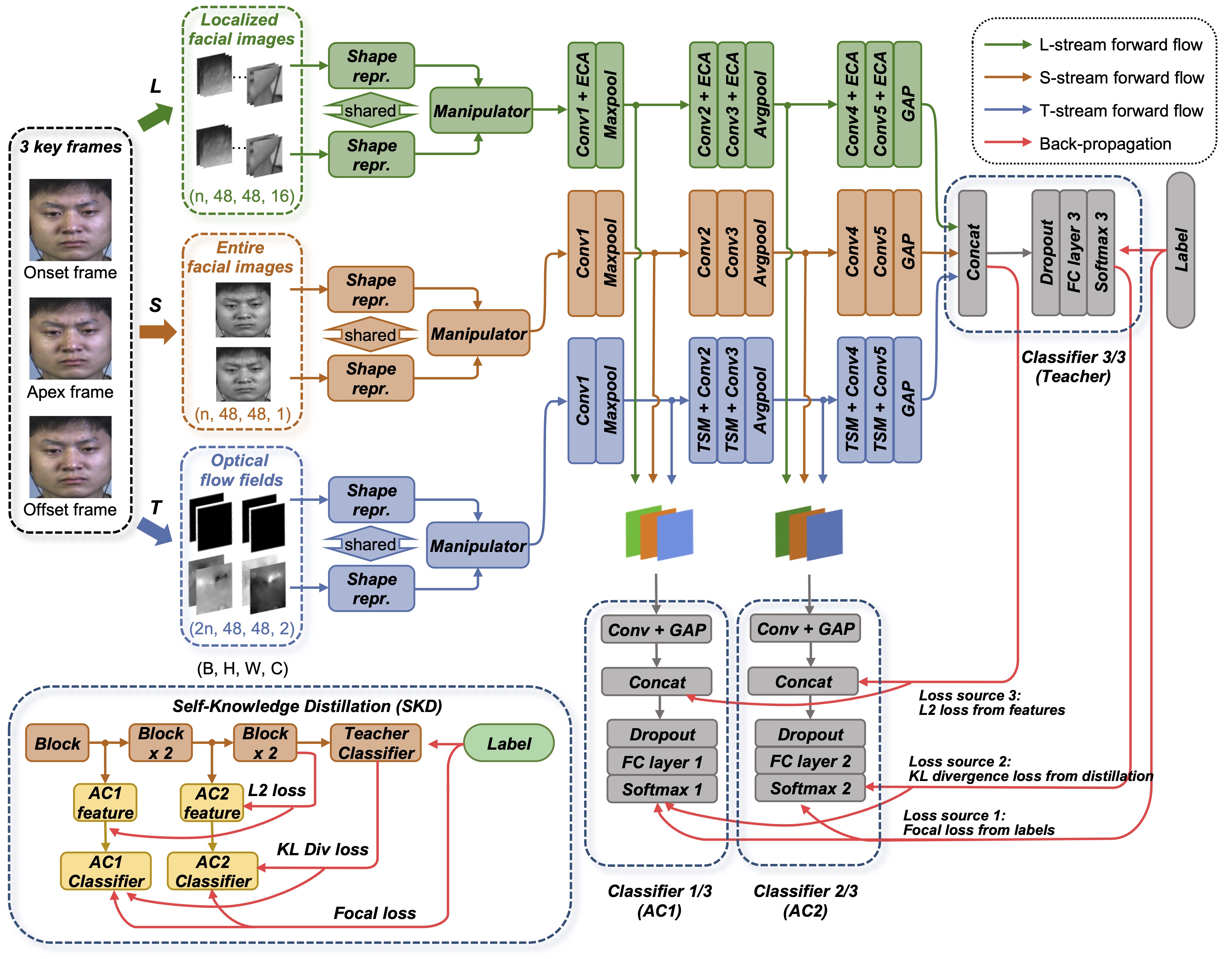}
\caption{The overview of our SKD-TSTSAN framework, including three streams with different types of input data. Each stream contains a motion magnification module. The L-stream includes ECA modules, and the T-stream contains TSMs. AC1 (Classifier 1/3) is added after the first convolutional block and AC2 (Classifier 2/3) is added after the third convolutional block for SKD; the deepest classifier (Classifier 3/3) serves as the teacher.}\label{fig1}
\end{figure}

The framework of our SKD-STSTAN is illustrated in Fig. \ref{fig1}. Firstly, the onset, apex, and offset frames of ME sequences as well as the motionless and real optical flow images are fed into the motion magnification modules to obtain the amplified shape representation, which is discussed in detail in Section \ref{subsec3-1}. Then, the magnified apex frame representation is input into the static-spatial stream (S-stream) to obtain the contour and appearance information of the whole face. The representation of the magnified local blocks is input into the local-spatial stream (L-stream) for the local ME feature extraction. The efficient channel attention (ECA) module \citep{wang2020eca} used in the L-stream is introduced in Section \ref{subsec3-2}. The amplified optical flow maps are input into the dynamic-temporal stream (T-stream), and the temporal shift module (TSM) \citep{lin2019tsm} used in the T-stream is introduced in Section \ref{subsec3-3}. Finally, the outputs of three recognition streams are fused into a fully connected layer to get the classification results. Furthermore, we add multiple auxiliary classifiers (ACs) in the network, leveraging SKD to enable
all blocks to fully learn the features of the training set, which is discussed in Section \ref{subsec3-4}.

\subsection{Motion Magnification Module}
\label{subsec3-1}

Due to the low intensity of ME, we enhance muscle movements using the motion magnification network proposed by \cite{oh2018learning}. To achieve an end-to-end framework and avoid noise in magnified images reconstructed in the decoder, our motion magnification module does not use a decoder. Given that each stream in our three-stream network has different inputs, we add separate motion magnification modules for each stream. Specifically, the input to the motion magnification network includes two frames with small motion displacement. For an ME sequence, the onset frame is the first frame where the ME begins, and the apex frame is the frame where the facial expression reaches maximum intensity, conveying significant emotional information. Therefore, the onset and apex frames are employed as inputs for the motion magnification module of the S-stream, and the concatenation of their $4\times 4$ grid divisions (16 patches per frame, $n=4$) as inputs for the L-stream’s motion magnification module. The choice of $n=4$ is validated by an ablation study on grid partition granularity reported in Section~\ref{subsubsec4-5-partition}. Particularly, since the T-stream extracts temporal information from optical flow fields, we utilize the optical flow corresponding to no motion and the optical flow between the onset and apex frames (and: between the apex and offset frames) as inputs for the motion magnification module in the T-stream.

The shape representation extractor used in our module is the encoder of the motion magnification network \citep{oh2018learning}, which maps each input frame through a series of convolutional layers into a latent shape representation that captures geometric boundary information. Such shape representations are more discriminative for modeling subtle muscle movement relationships than texture representations, which mainly encode color properties and provide little useful information for ME recognition \citep{lei2020novel}. Since the manipulator operates exclusively on shape representations—computing the motion-amplified output by differencing two shape representations and scaling by an amplification factor—we discard the texture branch and decoder, reducing both model complexity and reconstruction noise. The manipulator is formulated as:
\begin{equation}
{{G}_{m}}({{M}_{a}},{{M}_{b}},\alpha )={{M}_{a}}+h(\alpha \cdot g({{M}_{b}}-{{M}_{a}})),
\end{equation}
where ${{G}_{m}}(\cdot )$ denotes manipulator; ${{M}_{a}},{{M}_{b}}$ represent the shape representations of two input frames extracted by the encoder; $g(\cdot )$ is a $3\times 3$ convolution followed by ReLU; $\alpha $ denotes the amplification factor; $h(\cdot )$ represents a $3\times 3$ convolution followed by a $3\times 3$ residual block.

After the motion magnification modules, the motion-amplified face and optical flow field shape representations are obtained in each stream, and then we will extract global spatial information, local-spatial information, and temporal information in the S-, L-, and T-streams, respectively.

\subsection{Efficient Channel Attention Module}
\label{subsec3-2}

MEs only appear as subtle muscle movements in localized facial areas, such as the mouth and eyebrows. Therefore, the input to the L-stream, i.e., the $4\times 4$ facial blocks (16 patches), contains redundant information, which may affect emotion classification. However, we cannot simply select parts of the facial blocks, as different emotions correspond to distinct AUs involving various facial regions. For instance, AU6 (“Cheek Raiser”) mainly reflects happiness, whereas the ME for surprise does not involve this facial region. Thus, an adaptive approach should be used to extract important information from the $4\times 4$ facial blocks while suppressing unimportant information.

\begin{figure}[t]
\centering
\includegraphics{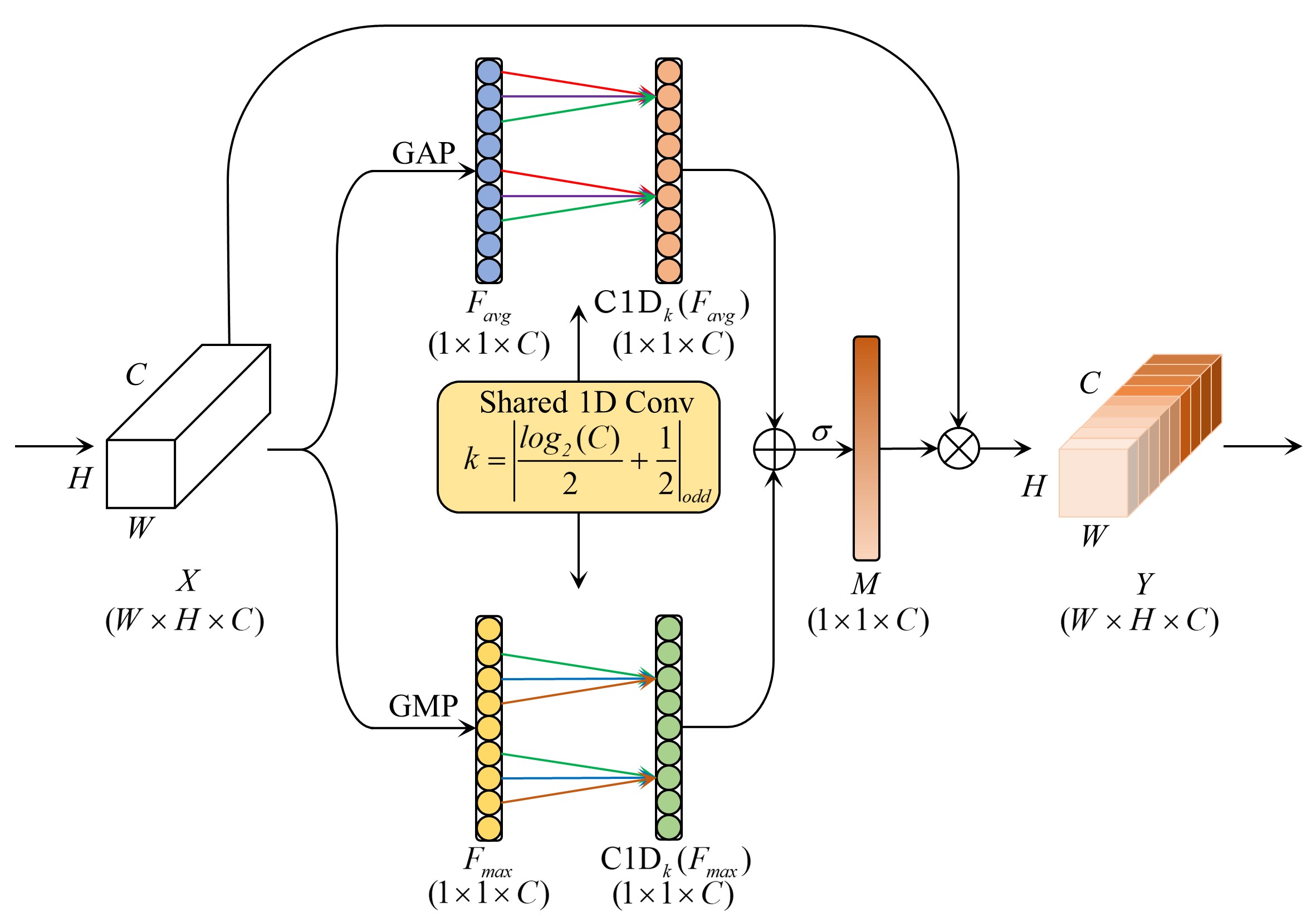}
\caption{Diagram of our enhanced efficient channel attention (ECA) module.}\label{fig2}
\end{figure}

Therefore, we employ efficient channel attention (ECA) modules \citep{wang2020eca} in the L-stream, introducing an efficient channel attention mechanism. Specifically, compared to the original squeeze-and-excitation (SE) block \citep{hu2018squeeze}, the ECA module avoids disrupting the direct relationship between channels and their weights and uses 1D convolution to achieve local cross-channel interaction. As illustrated in Fig. \ref{fig2}, the input feature map is $X\in {{\mathbb{R}}^{W\times H\times C}}$, where $W,H,C$  represents the width, height, and channel dimensions. The original ECA module employs global average pooling (GAP) to aggregate global spatial features into ${{F}_{avg}}\in {{\mathbb{R}}^{1\times 1\times C}}$. Additionally, we employ global maximum pooling (GMP) to provide the most significant information for each channel, obtaining ${{F}_{max}}\in {{\mathbb{R}}^{1\times 1\times C}}$. Then, by performing a 1D convolution with a kernel size of $k$ on ${F}_{avg}$ and ${F}_{max}$, we capture cross-channel interactions and obtain the channel weights $M\in {{\mathbb{R}}^{1\times 1\times C}}$, which are represented as follows:
\begin{equation}
M=\sigma (\text{C1}{{\text{D}}_{k}}({{F}_{avg}})+\text{C1}{{\text{D}}_{k}}({{F}_{max}})),
\end{equation}
where $\text{C1}{{\text{D}}_{k}}$ denotes the 1D convolution with kernel size $k$, which is adaptively adjusted based on the number of channels:
\begin{equation}
k={{\left| \frac{lo{{g}_{2}}(C)}{2}+\frac{1}{2} \right|}_{odd}},
\end{equation}
where ${{\left| t \right|}_{odd}}$ denotes the nearest odd value of $t$. Finally, the channel-wise refined output feature map is calculated as follows:
\begin{equation}
Y=M\otimes X,
\end{equation}
where $\otimes $ represents element-wise multiplication.

We added the enhanced ECA module after each convolutional layer in the L-stream, using the channel attention mechanism to suppress the impact of redundant information on emotion classification.

\begin{figure}[t]
\centering
\includegraphics[width=\linewidth]{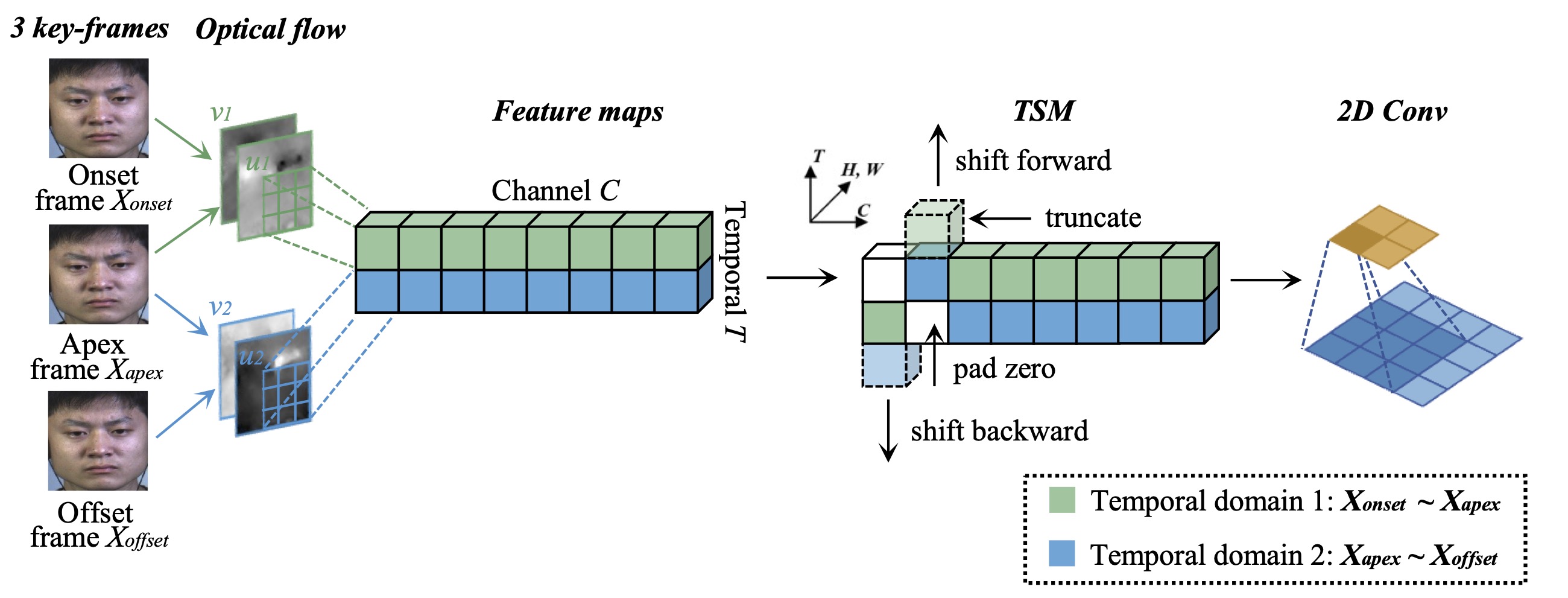}
\caption{Illustration of the bi-directional TSM used in T-stream. The optical flow fields in two different temporal domains are calculated, where $u$ denotes the horizontal optical flow and $v$ represents the vertical optical flow. Temporal modeling is achieved by shifting channels along the temporal dimension. The subsequent 2D convolution performs the multiply-accumulate operation on the features from different frames’ channels.}\label{fig3}
\end{figure}

\subsection{Temporal Shift Module}
\label{subsec3-3}

3D CNNs can effectively mine spatiotemporal information but are computationally intensive. In order to extract spatiotemporal information while keeping the complexity of 2D CNNs, we incorporate the temporal shift module (TSM) \citep{lin2019tsm} into the T-stream.

The schematic of the TSM we use is shown in Fig. \ref{fig3}. Since the ME videos are offline videos, with access to future frames, the bi-directional TSM is employed. For a tensor with C channels and T frames, we use varying colors to represent features at different time stamps. By shifting part of the channels forward by one frame and another part backward by one frame along the temporal dimension, we achieve the mixing of information from past and
\begin{figure}[t]
\centering
\includegraphics[scale=1.35]{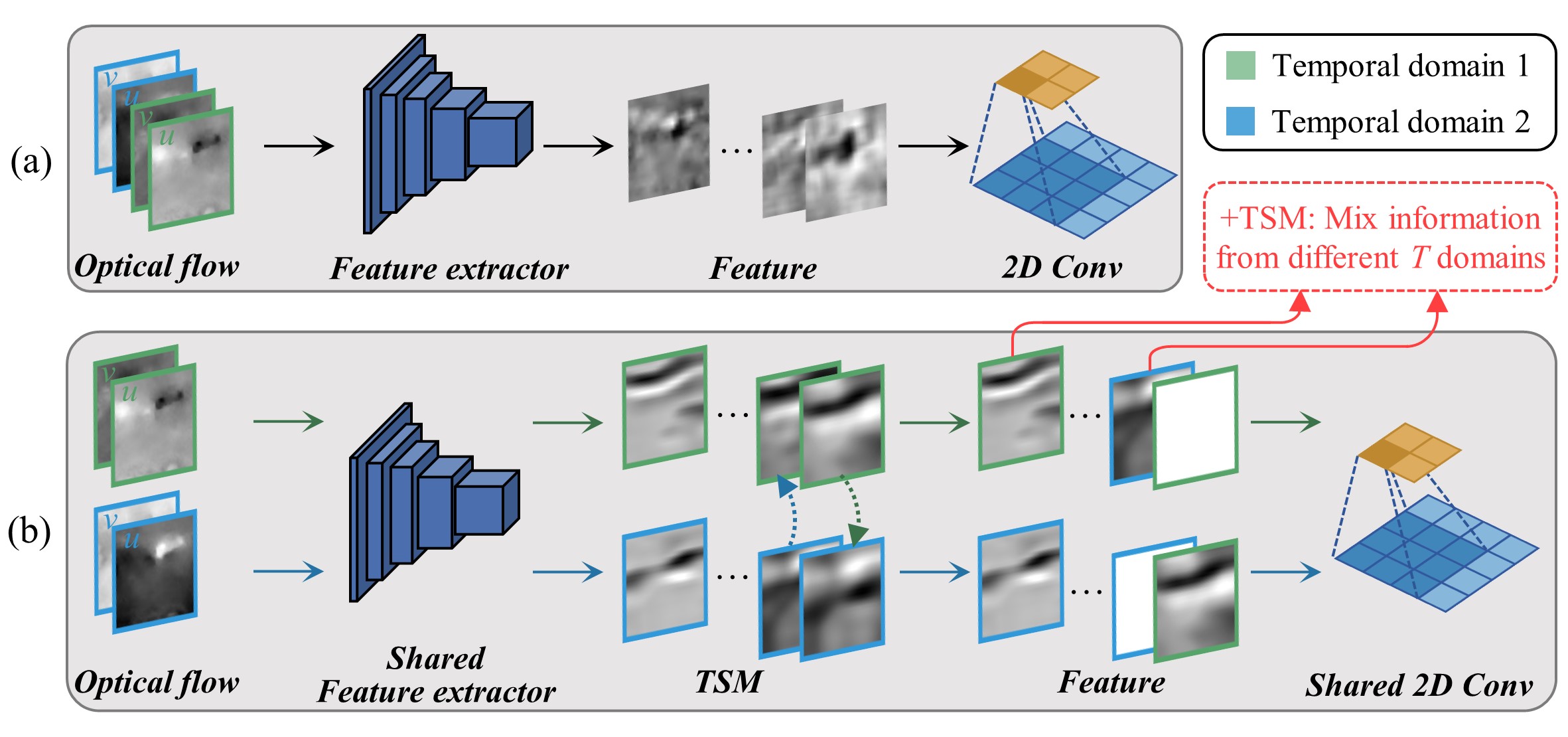}
\caption{The forward propagation of the T-stream in the network. (a) and (b) represent the forward propagation of the T-stream in the network without and with TSM, respectively. Different border colors indicate that the information comes from different temporal domains.}\label{fig4}
\end{figure}
future frames with the current frame. The excess channels are truncated and the blank frame parts are filled with zeros. The shift operation in TSM does not require any multiplication, and the multiply-accumulate operation of features from different frames’ channels is completed by the subsequent 2D convolution. Therefore, efficient temporal modeling is achieved at no extra cost.

As shown in Fig. \ref{fig4}(a), in the original TSCNN, the input to the T-stream is the concatenation of horizontal and vertical optical flows between the onset frame, apex frame, and offset frame pairs. It can be seen that 2D CNNs without TSM have difficulty in capturing temporal relationships in forward propagation. To introduce TSM, we reorganize the T-stream inputs along the temporal dimension: the onset-to-apex optical flow (horizontal component $u_1$ and vertical component $v_1$) and the apex-to-offset optical flow ($u_2$, $v_2$) form two temporal domains, each represented as a 2-channel map of size $48\times48$. The two maps are stacked and treated as a 2-frame temporal sequence ($t\!=\!0$: onset-to-apex; $t\!=\!1$: apex-to-offset), enabling TSM to exchange information across them.

As illustrated in Fig. \ref{fig4}(b), TSMs are inserted before all convolutional layers in the intermediate stages of the T-stream. The shift ratio must be balanced: too large a ratio degrades spatial feature learning, while too small a ratio limits temporal interaction \citep{lin2019tsm}. Therefore, the first $1/8$ of channels at $t\!=\!0$ receive features from $t\!=\!1$ (apex-to-offset information propagates backward to the onset-to-apex domain), the following $1/8$ at $t\!=\!1$ receive features from $t\!=\!0$ (onset-to-apex information propagates forward), and the remaining $3/4$ of channels are left unchanged. The tensor dimensionality is preserved through each shift, and no additional parameters are introduced; the subsequent 2D convolution performs the cross-temporal feature fusion.

\subsection{Self-Knowledge Distillation}
\label{subsec3-4}

In order to further improve our lightweight three-stream network's performance, we introduce an effective model compression strategy, self-knowledge distillation (SKD) \citep{zhang2019your}. In traditional knowledge distillation, a complex teacher model is first trained, and then a compact student model imitates the predictions of the teacher model \citep{hinton2015distilling}. However, it is difficult to design an effective over-parameterized model to act as the teacher model on a small-scale ME dataset \citep{zhang2019your}. Instead, in SKD, the model acts as its own teacher, using knowledge acquired from the training set to guide its own training. This approach requires less training time than the two-stage method and achieves better results.

We adopt the Be Your Own Teacher (BYOT) framework proposed by \cite{zhang2019your}, adding two auxiliary classifiers (ACs) after the first and third convolutional blocks of the network. Specifically, AC1 (Classifier 1/3) is attached after the first convolutional block and acts as the shallowest student classifier; AC2 (Classifier 2/3) is attached after the third convolutional block and acts as the intermediate student classifier. The final classifier at the end of the full network (Classifier 3/3) serves as the teacher in SKD. Each AC has a three-stream architecture similar to the main network, including convolutional layers and a fully connected layer, and contains the previously mentioned ECA modules and TSMs. The ACs are only used during training and can be removed during inference. During the training stage, the deepest part of the network acts as the teacher model, while the shallow parts with the corresponding classifiers serve as the student model. We employ three kinds of loss functions: focal loss (FL) \citep{lin2017focal}, Kullback-Leibler (KL) divergence loss, and L2 loss. Then the computation methods for each loss function will be detailed.

Given a training set $D$ for a MER task, we use $\{x,y\}\in D$ to represent each sample in the dataset, where $x$ denotes the input and $y$ denotes the label. Suppose there are $K$ classes, so $y\in\{1,\ldots,K\}$. The multiple classifiers in the network are represented as $\Theta =\{{{\theta }_{i}}\}_{i=1}^{C}$, where $C$ indicates the number of classifiers, and ${{\theta }_{C}}$ represents the deepest classifier. We use ${{z}^{c}}$ to denote the output following the fully connected layer of classifier ${{\theta }_{c}}$, and $z_{i}^{c}$ to denote the logit corresponding to the ${{i}_{th}}$ class in the output ${{z}^{c}}$. Each classifier is followed by a softmax layer:
\begin{equation}
q_{i}^{c}(T)=\frac{\exp (z_{i}^{c}/T)}{\sum\limits_{j=1}^{K}{\exp (z_{j}^{c}/T)}},
\end{equation}
where $q_{i}^{c}$ indicates the class probability of the ${{i}_{th}}$ class for the classifier ${{\theta }_{c}}$. $T$ denotes the distillation temperature, where a large $T$ is used to soften the class probability distribution.

All classifiers are supervised by the labels. To address data imbalance and focus more on hard examples, we choose focal loss to implement the supervision from labels. We first define ${t_{i}}$ to represent the true probability distribution as follows:
\begin{equation}
{{t}_{i}}=\left\{ \begin{array}{*{35}{l}}
   1, & i=true\ label  \\
   0, & i\ne true\ label  \\
\end{array} \right..
\end{equation}

The multi-class focal loss we use is defined as follows:
\begin{equation}
L_{FL}^{i}({{q}^{i}}(T),y)=-\sum\limits_{k=1}^{K}{{{\alpha }_{k}}{{(1-q_{k}^{i}(T))}^{\gamma }}{{t}_{k}}\log (q_{k}^{i}(T))},
\end{equation}
where ${{q}^{i}}$ denotes the output probability of the classifier ${{\theta }_{i}}$. The focusing parameter $\gamma $ determines the rate at which the weights of easy samples are reduced, and ${{\alpha }_{k}}$ represents the weighting factor for the class ${{k}_{th}}$. ${{\alpha }_{k}}$ is used to address class imbalance, generally set utilizing the inverse class frequency.

The supervision for the deepest classifier ${{\theta }_{C}}$ comes only from ${{L}_{FL}}$, while the supervision for the auxiliary classifiers also comes from the KL divergence loss ${{L}_{KL}}$ and the L2 loss ${{L}_{L2}}$. ${{L}_{KL}}$ is calculated between ${{q}^{i}}$ and the probability output ${{q}^{C}}$ of the deepest classifier, reflecting the supervision from distillation. ${{L}_{KL}}$ is represented as follows:
\begin{equation}
L_{KL}^{i}({{q}^{i}}(T),{{q}^{C}}(T))={{T}^{2}}\sum\limits_{k=1}^{K}{q_{k}^{C}(T)\log \frac{q_{k}^{C}(T)}{q_{k}^{i}(T)}}.
\end{equation}

The purpose of ${{L}_{L2}}$ is to use the output of hidden layers of the deepest classifier to guide the learning of the shallow classifiers, aiming to reduce the distance between their feature maps. Since feature maps at different depths have different sizes, convolutional layers are employed in ACs to align them. ${{L}_{L2}}$ is defined as follows:
\begin{equation}
L_{L2}^{i}=\left\| {{F}_{i}}-{{F}_{C}} \right\|_{2}^{2},
\end{equation}
where ${{F}_{i}}$ and ${{F}_{C}}$ represent the features of the classifier ${{\theta }_{i}}$ and the deepest classifier ${{\theta }_{C}}$, respectively.

In summary, the total loss function of the proposed SKD-TSTSAN is expressed as:
\begin{equation}
L=\sum\limits_{i=1}^{C}{[(1-{{\lambda }_{1}})L_{FL}^{i}({{q}^{i}}(1),y)+{{\lambda }_{1}}\cdot L_{KL}^{i}({{q}^{i}}(T),{{q}^{C}}(T))+{{\lambda }_{2}}\cdot L_{L2}^{i}]},
\end{equation}
where ${{\lambda }_{1}}$ and ${{\lambda }_{2}}$ are hyperparameters used to balance the loss terms. For the deepest classifier, both ${{\lambda }_{1}}$ and ${{\lambda }_{2}}$ are set to 0, indicating that its supervision comes only from the labels.

\section{Experiments}
\label{sec4}
In this section, we will provide the details of our experiments, including the datasets used, data pre-processing, implementation details, and evaluation metrics. Additionally, the results and analysis of the comparative experiments with other SOTA methods, ablation studies, and parameter setting experiments will be provided.

\subsection{Datasets}
\label{subsec4-1}
We evaluate our SKD-TSTSAN on five spontaneous ME datasets: the CASME II \citep{yan2014casme}, SAMM \citep{davison2016samm}, SMIC \citep{li2013spontaneous}, MMEW \citep{ben2021video}, and CAS(ME)$^3$ \citep{li2022cas} databases. The details of these datasets are introduced below.
\begin{enumerate} [(1)]
    \item The CASME II dataset includes 255 ME samples from 26 participants, which are recorded at 200 fps with a facial resolution of $280\times 340$. These samples are divided into seven categories: Happiness (32), Disgust (61), Repression (27), Surprise (25), Sadness (7), Fear (2), and Others (99), where the numbers in brackets represent the number of corresponding MEs. Due to the small number of samples in the “Sadness” and “Fear” categories, each having fewer than 10 samples, the CASME II dataset is usually used to study a 5-category MER task involving the remaining 5 categories. Additionally, some studies classify emotion categories into 3 categories: the “Positive” emotion category contains “Happiness” class, the “Negative” category contains “repression” and “disgust” classes, and the “Surprise” category includes only the “Surprise” class.
    \item The SAMM dataset includes 159 ME samples from 29 participants, recorded at a frame rate of 200 fps and a facial resolution of $400\times 400$. Eight emotion categories are labeled in SAMM: Anger (57), Happiness (26), Surprise (15), Contempt (12), Disgust (9), Fear (8), Sadness (6), and Others (26). This dataset is typically divided into three categories: a "Positive" category, which includes the "Happiness" class; a "Negative" category, comprising the "Anger", "Disgust", "Contempt", "Sadness", and "Fear" classes; and a "Surprise" category, which consists solely of the "Surprise" class.
    \item The SMIC dataset provides three subsets recorded with different types of cameras: HS (high-speed camera), VIS (visible light camera), and NIR (near-infrared camera). In our experiments, we adopted the HS subset, which was collected using a high-speed camera at a frame rate of 100 fps. This subset consists of 164 ME samples from 16 participants, with a facial resolution of $190\times 230$. The emotion categories in SMIC are labeled as Negative (70), Positive (51), and Surprise (43). Since the SMIC dataset does not provide labels for apex frames, we utilized the apex frame labels provided by the CapsuleNet method \citep{van2019capsulenet}.
    \item The MMEW dataset includes 300 ME samples and 900 macro-expression samples from 36 participants. They are recorded using a frame rate of 90 fps with a facial resolution of $400\times 400$. The ME samples are divided into seven categories: Happiness (36), Disgust (72), Surprise (89), Fear (16), Sadness (13), Anger (8) and Others (66). The MMEW dataset is typically divided into four and six emotion categories. For the 6-class MER task, the "Anger" category is excluded. In the 4-class MER task, four categories with similar sample sizes are selected: Happiness, Disgust, Surprise, and Others.
    \item The CAS(ME)$^3$ dataset is a large-scale spontaneous ME dataset, which provides about 80 hours of video at a frame rate of 30 fps and offers depth information as an additional modality. According to the latest version of the dataset annotations, there are 860 manually labeled MEs, divided into seven categories: Anger (64), Disgust (250), Fear (86), Happy (55), Others (161), Sad (57), and Surprise (187). This dataset is typically divided into 3, 4, and 7 emotion categories. For the 3-category classification, negative emotions such as "Anger", "Disgust", "Fear", and "Sad" are grouped into a "Negative" category, "Happy" is assigned to a "Positive" category, and "Surprise" forms the third category. In the 4-category classification, an additional "Others" category is included. The 7-category classification utilizes all the official emotion categories.
    \item The DFME dataset \citep{zhao2023dfme} is a large-scale spontaneous ME dataset designed for dynamic facial micro-expression recognition research. It comprises 2,629 ME samples from 259 subjects, recorded at frame rates of 200, 300, or 500 fps. The samples are labeled with seven emotion categories: Anger (241), Contempt (171), Disgust (735), Fear (365), Happiness (311), Sadness (359), and Surprise (447). The dataset provides an official split: a training set (1,856 samples from 176 subjects), Test A (474 samples from 42 subjects), and Test B (299 samples from 41 subjects). In our experiments, we train on the full training set and evaluate on both test sets to assess the generalization capability of our model on this large-scale, diverse benchmark.
\end{enumerate}

To comprehensively evaluate the effectiveness of our SKD-TSTSAN in the field of MER, we conducted both composite dataset evaluation \citep{see2019megc} and single dataset evaluation in our experiments. For the composite dataset evaluation, CASME II, SAMM, and SMIC were merged into a 3-category composite dataset, where the MER task utilized the shared emotion labels across these datasets: Positive (109 samples), Negative (250 samples), and Surprise (83 samples). In the single dataset evaluation, experiments were performed on individual datasets. Specifically, we conducted 5-class MER on the CASME II dataset, 4-class and 6-class MER on the MMEW dataset, 3-class, 4-class, and 7-class MER on the CAS(ME)$^3$ dataset, and 7-class MER on the DFME dataset.

\subsection{Data Pre-Processing and Implementation Details}
\label{subsec4-2}
As the proposed SKD-TSTSAN requires knowledge transfer from macro-expression data, we first pre-train it on the CK+ dataset \citep{lucey2010extended}, which includes seven expressions: Happiness, Anger, Disgust, Contempt, Fear, Sadness, and Surprise. The CK+ dataset, being a macro-expression dataset, lacks annotations like the apex frames present in ME datasets. Therefore, to pre-train the network using this dataset, we designate the last frame with the most obvious muscle movement of each sample as the apex frame, and the first frame as both the onset and offset frames. After pre-training, we employ transfer learning to transfer the knowledge learned from the macro-expression dataset to the MER task. Specifically, we remove the final linear layer and use the remaining weights as the initial weights for training on the micro-expression dataset. By transferring knowledge from macro-expressions to micro-expressions, the model benefits from the rich information in macro-expression data while adapting to the subtle and rapid dynamics of micro-expressions.

For each input sample, Retinaface \citep{deng2019retinaface} is first employed to detect the facial region and resize it to $128\times 128$. Then, we compute optical flow features using the TV-L1 method \citep{zach2007duality}. The entire facial region is scaled down to $48\times 48$ and fed into the S-stream of our network. The $128\times 128$ facial region is divided into a $4\times 4$ grid, resulting in 16 patches, each scaled to $48\times 48$. These patches are concatenated along the channel dimension and fed into the L-stream. Finally, the horizontal and vertical optical flow fields between the three key-frames are combined into two $48\times 48$ 2-channel images, which are fed into the T-stream of the network.

We implement our SKD-TSTSAN on an NVIDIA-RTX-4090 GPU. The Adam optimizer is utilized with learning rate, betas and weight decay set to 1e-3, (0.9, 0.99) , 5e-4, respectively. Loss weights ${{\lambda }_{1}}=0.1$, ${{\lambda }_{2}}=1e-6$ and distillation temperature $T=3$ are chosen as default values. In the multi-class focal loss, the focusing parameter $\gamma $ is set to 2, and the weighting factor ${{\alpha }_{k}}$ is positively correlated with the inverse of the sample size for each category. In our motion magnification module, we set the amplification factor to 2. During the pre-training stage on the CK+ dataset, we configure the training iterations to 10k and the batch size to 16. When training on the CASME II, MMEW, and composite datasets, the batch size is set to 16, while for the CAS(ME)$^3$ and DFME datasets, it is set to 32. Training iterations on the ME dataset are set to 20k and early stopping is used to alleviate overfitting.

\subsection{Evaluation Metrics}
\label{subsec4-3}
For all experiments on the five public datasets, we employ leave-one-subject-out (LOSO) cross-validation. The unweighted accuracy (UAR) and unweighted F1 score (UF1) are used to evaluate the MER results. For convenience, $T{{P}_{i}}$, $F{{P}_{i}}$, and $F{{N}_{i}}$ are used to represent the predicted numbers of true positives, false positives, and false negatives for each class. $UF1$ and $UAR$ are calculated as follows:
\begin{equation}
UF1=\frac{1}{C}\sum\limits_{i=1}^{C}{\frac{2\times T{{P}_{i}}}{T{{P}_{i}}+F{{P}_{i}}+F{{N}_{i}}}},
\end{equation}
\begin{equation}
UAR=\frac{1}{C}\sum\limits_{i=1}^{C}{\frac{T{{P}_{i}}}{{{N}_{i}}}},
\end{equation}
where $C$ represents the number of emotion categories in the dataset, and ${{N}_{i}}$ denotes the total number of MEs in the ${{i}_{th}}$ class.

\begin{table}[t]
\tiny
\setlength{\abovecaptionskip}{0.1cm}
\centering
\caption{Comparisons with other approaches on composite (Full), CASME II, SAMM, and SMIC datasets.}\label{table1}
\begin{tabular}{ccccccccc}
\hline\hline
\multirow{2}{*}{Method} & \multicolumn{2}{c}{Full} & \multicolumn{2}{c}{CASME II} & \multicolumn{2}{c}{SAMM} & \multicolumn{2}{c}{SMIC} \\
                        & UF1 & UAR & UF1 & UAR & UF1 & UAR & UF1 & UAR \\ \hline
OFF-ApexNet \citep{gan2019off}             & 71.96    & 70.96    & 87.64    & 86.81    & 54.09    & 53.92    & 68.17    & 66.95    \\
STSTNet \citep{liong2019shallow}                 & 73.53    & 76.05    & 83.82    & 86.86    & 65.88    & 68.10    & 68.01    & 70.13    \\
CapsuleNet \citep{van2019capsulenet}              & 65.20    & 65.06    & 70.68    & 70.18    & 62.09    & 59.89    & 58.20    & 58.77    \\
Dual-Inception \citep{zhou2019dual}         & 73.22    & 72.78    & 86.21    & 85.60    & 58.68    & 56.63    & 66.45    & 67.26    \\
EMR \citep{liu2019neural}                     & 78.85    & 78.24    & 82.93    & 82.09    & 77.54    & 71.52    & 74.61    & 75.30    \\
FeatRef \citep{zhou2022feature}                & 78.38    & 78.32    & 89.15    & 88.73    & 73.72    & 71.55    & 70.11    & 70.83    \\
HTNet \citep{wang2024htnet}                  & 86.03    & 84.75    & 95.32    & 95.16    & 81.31    & 81.24    & 80.49    & 79.05    \\
CSARNet \citep{zhao2025channel}             & 82.39    & 83.00    & 92.54    & 92.98    & 78.94    & 79.24    & 76.05    & 76.39    \\
MPFNet \citep{ma2025mpfnet}                & 84.00    & 84.60    & 91.10    & 92.30    & 79.50    & 83.90    & 80.60    & 80.90    \\
HFA-Net \citep{zhang2025hfa}               & 88.00    & 86.60    & 97.38    & 97.54    & \textbf{90.02} & \textbf{89.38} & 77.86    & 77.86    \\
CausalNet \citep{zhang2025causalnet}       & \underline{89.30} & \underline{89.41} & \underline{97.48} & \underline{97.82} & 87.08    & \underline{85.22} & \underline{84.05} & \underline{84.33} \\
\textbf{SKD-TSTSAN (ours)} & \textbf{90.20} & \textbf{90.83} & \textbf{98.58} & \textbf{98.58} & \underline{87.70} & 84.00    & \textbf{84.16} & \textbf{85.71} \\ \hline\hline
\end{tabular}
\end{table}

\begin{table}[t]
\footnotesize
\setlength{\abovecaptionskip}{0.1cm}
\begin{minipage}{1.0\textwidth}
\centering
\caption{Comparisons with other approaches on CASME II dataset.}\label{table2}
\begin{tabular}{p{6.5cm}<{\centering} p{1.5cm}<{\centering} p{2cm}<{\centering}p{2cm}<{\centering}}
\hline\hline
Method            & \# Classes & UF1 (\%) & UAR (\%) \\ \hline
LR-GACNN \citep{kumar2021micro}          & 5          & 70.90     & 81.30     \\
AMAN \citep{wei2022novel}              & 5          & 71.00       & 75.40     \\
Graph-TCN \citep{lei2020novel}         & 5          & 72.46    & 73.98    \\
MiMaNet \citep{xia2021micro}           & 5          & 75.90     & 79.90     \\
SMA-STN \citep{liu2020sma}           & 5          & 79.46    & 82.59    \\
TSCNN \citep{song2019recognizing}             & 5          & 80.70     & 80.97    \\
MER-CLIP \citep{liu2025mer}            & 5          & 83.78    & N/A    \\
$\mu $-BERT \citep{nguyen2023micron}            & 5          & \underline{85.53}    & \underline{83.48}    \\
\textbf{SKD-TSTSAN (ours)} & 5          & \textbf{90.34}    & \textbf{89.47}    \\ \hline\hline
\end{tabular}
\end{minipage}
\end{table}

\subsection{Comparative Experiments Results}
\label{subsec4-4}
\begin{figure}[t]
\centering
\includegraphics[scale=0.65]{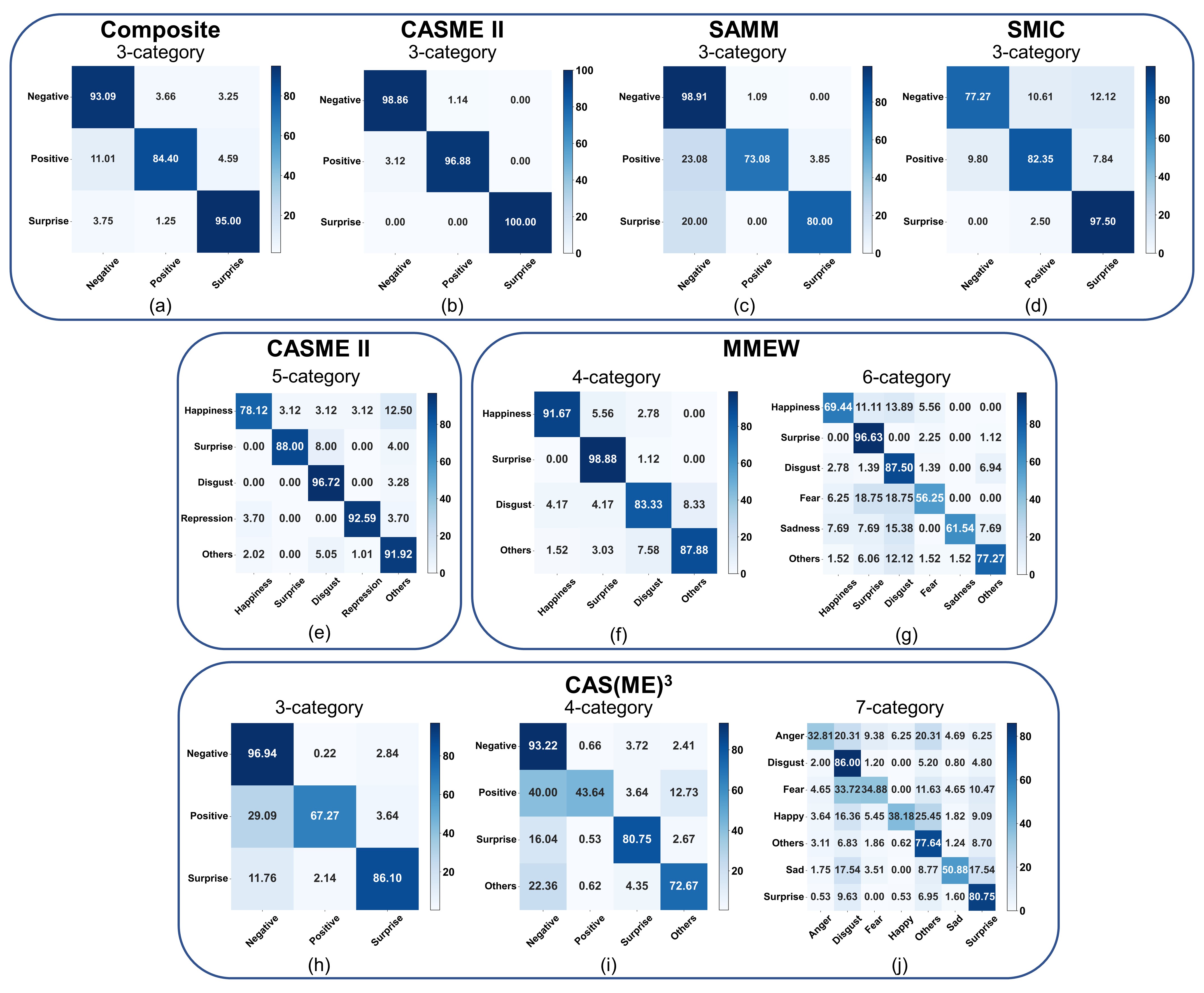}
\caption{Confusion matrices for MER with SKD-TSTSAN on the composite, CASME II, SAMM, SMIC, MMEW, and CAS(ME)$^3$ datasets. (a), (b), (c), and (d) represent the 3-class classification results on the full composite dataset, CASME II, SAMM, and SMIC datasets, respectively. (e) denotes the 5-class classification results on CASME II. (f) and (g) indicate the 4-class and 6-class classification results on MMEW. (h), (i), and (j) show the 3-class, 4-class, and 7-class classification results on CAS(ME)$^3$, respectively.}\label{fig5}
\end{figure}

In this section, we compare the results of our method with other SOTA methods using five public ME datasets (CASME II, SAMM, SMIC, MMEW, and CAS(ME)$^3$). As seen in Tables \ref{table1} to \ref{table4}, our SKD-TSTSAN demonstrates significant performance improvements in the MER task. Furthermore, the confusion matrices of our SKD-TSTSAN on different datasets are shown in Fig. \ref{fig5}. For all comparison methods, we include results only when the reported dataset setting, evaluation protocol, and metrics are directly comparable to ours, or when we can reproduce the method under our experimental setting. Because directly comparable published results are limited for the 4-class and 6-class MMEW settings, we reproduce several representative methods, including CapsuleNet \citep{van2019capsulenet}, MMNet \citep{li2022mmnet}, STSTNet \citep{liong2019shallow}, and TSCNN \citep{song2019recognizing}. For the other datasets, the comparison results are taken from the original publications.

Table \ref{table1} presents the quantitative comparison results for the 3-category MER task on the composite (Full), CASME II, SAMM, and SMIC datasets. To provide a comprehensive evaluation against recent advances, we include several recently published SOTA methods: CSARNet \citep{zhao2025channel}, MPFNet \citep{ma2025mpfnet}, HFA-Net \citep{zhang2025hfa}, and CausalNet \citep{zhang2025causalnet}. Our SKD-TSTSAN achieves the best performance on the composite dataset, with a UF1 of 90.20\% and a UAR of 90.83\%, surpassing the second-best CausalNet \citep{zhang2025causalnet} by 0.90\% in UF1 and 1.42\% in UAR. On the CASME II dataset, SKD-TSTSAN also achieves the highest results with a UF1 and UAR of 98.58\%, outperforming CausalNet (97.48\% UF1 and 97.82\% UAR) by 1.10\% and 0.76\%, respectively. On the SAMM dataset, HFA-Net \citep{zhang2025hfa} achieves the best UF1 of 90.02\% and UAR of 89.38\%, slightly surpassing SKD-TSTSAN (87.70\% UF1 and 84.00\% UAR), which ranks second in UF1. On the SMIC dataset, SKD-TSTSAN achieves the highest UF1 of 84.16\% and UAR of 85.71\%, exceeding CausalNet (84.05\% UF1 and 84.33\% UAR). Overall, SKD-TSTSAN demonstrates superior performance on most benchmarks and strong generalization ability across different datasets. Most methods are deep learning approaches that utilize optical flow features. The key difference is that our SKD-TSTSAN uses the optical flow fields between the three key frames (onset, apex, and offset) as the input to the T-stream, while the other methods \citep{liong2019shallow, wang2024htnet, gan2019off, liu2019neural, zhou2022feature} only utilize the optical flow field between the onset frame and the apex frame. By leveraging motion information from different temporal regions, SKD-TSTSAN fully considers the temporal dynamics of micro-expressions, which achieves superior recognition performance.

As shown in Table \ref{table2}, we report the results from the single dataset evaluation on CASME II. For the 5-category emotion classification, SKD-TSTSAN achieved the highest UF1 of 90.34\% and UAR of 89.47\%, compared to $\mu$-BERT's 85.53\% UF1 and 83.48\% UAR. MER-CLIP reports a UF1 of 83.78\% under the same 5-class setting, while UAR is not available in the original report. Furthermore, compared to our baseline TSCNN, SKD-TSTSAN shows a 9.64\% UF1 and 8.50\% UAR improvement, proving the effectiveness of our submodules.

\begin{table}[t]
\footnotesize 
\setlength{\abovecaptionskip}{0.1cm}
\begin{minipage}{1.0\textwidth}
\centering
\caption{Comparisons with other approaches on MMEW dataset.}\label{table3}
\begin{tabular}{p{6.5cm}<{\centering} p{1.5cm}<{\centering} p{2cm}<{\centering}p{2cm}<{\centering}}
\hline\hline
Method            & \# Classes & UF1 (\%) & UAR (\%) \\ \hline
CapsuleNet \citep{van2019capsulenet}          & 6          & 44.13     & 45.31     \\
MMNet \citep{li2022mmnet}              & 6          & 57.05       & 56.95     \\
STSTNet \citep{liong2019shallow}         & 6          & 64.13    & 62.65    \\
TSCNN \citep{song2019recognizing}             & 6          & \underline{70.43}       & \underline{68.76}       \\
\textbf{SKD-TSTSAN (ours)} & 6          & \textbf{77.11}    & \textbf{74.77}    \\ \hline
CapsuleNet \citep{van2019capsulenet}           & 4          & 64.02    & 63.15    \\
STSTNet \citep{liong2019shallow}       & 4          & 78.99    & 78.89    \\
TSCNN \citep{song2019recognizing}            & 4          & \underline{86.46}    & 86.09    \\
MMNet \citep{li2022mmnet}             & 4          & 86.35    & \underline{87.45}    \\
\textbf{SKD-TSTSAN (ours)} & 4          & \textbf{90.41}    & \textbf{90.44}    \\ \hline\hline
\end{tabular}
\end{minipage}
\end{table}

The performance of SKD-TSTSAN on the MMEW dataset is compared with different methods in Table \ref{table3}. To ensure direct comparability under our 4-class and 6-class MMEW settings, we reproduce CapsuleNet for both 4-class and 6-class tasks, MMNet for the 6-class task, STSTNet for both 4-class and 6-class tasks, and TSCNN for both 4-class and 6-class tasks. For MMNet’s performance on the 4-category MER task, the optimal performance reported in its original paper is provided. For six categories, CapsuleNet obtains the lowest UF1 and UAR since it only uses the apex frame as the network 
\begin{table}[t]
\footnotesize  
\setlength{\abovecaptionskip}{0.1cm}
\caption{Comparisons with other approaches on CAS(ME)$^3$ dataset.}\label{table4}
\begin{minipage}{1.0\textwidth}
\centering
\begin{tabular}{p{6.5cm}<{\centering} p{1.5cm}<{\centering} p{2cm}<{\centering}p{2cm}<{\centering}}
\hline\hline
Method            & \# Classes & UF1 (\%) & UAR (\%) \\ \hline
FR \citep{zhou2022feature}          & 3          & 34.93     & 34.13     \\
STSTNet \citep{liong2019shallow}              & 3          & 37.95       & 37.92     \\
RCN-A \citep{xia2020revealing}         & 3          & 39.28    & 38.93    \\
$\mu $-BERT \citep{nguyen2023micron}             & 3          & 56.04       & 61.25       \\
FED-PsyAU \citep{li2025fed}           & 3          & 62.21    & 62.26    \\
MER-CLIP \citep{liu2025mer}            & 3          & \underline{78.32}    & \underline{76.06}       \\
\textbf{SKD-TSTSAN (ours)} & 3          & \textbf{86.48}    & \textbf{83.44}    \\ \hline
AlexNet \citep{li2022cas}           & 4          & 29.15    & 29.10    \\
AlexNet(+Depth) \citep{li2022cas}       & 4          & 30.01    & 29.82    \\
SFAMNet \citep{liong2024sfamnet}            & 4          & 44.62    & 47.97    \\
$\mu $-BERT \citep{nguyen2023micron}             & 4          & 47.18    & 49.13    \\
MER-CLIP \citep{liu2025mer}            & 4          & \underline{65.44}    & \underline{62.42}       \\
\textbf{SKD-TSTSAN (ours)} & 4          & \textbf{76.40}    & \textbf{72.57}    \\ \hline
AlexNet \citep{li2022cas}           & 7          & 17.59    & 18.01    \\
AlexNet(+Depth) \citep{li2022cas}       & 7          & 17.73    & 18.29    \\
SFAMNet \citep{liong2024sfamnet}            & 7          & 23.65    & 23.73    \\
$\mu $-BERT \citep{nguyen2023micron}             & 7          & 32.64    & 32.54    \\
MER-CLIP \citep{liu2025mer}            & 7          & \underline{49.97}    & \underline{50.14}       \\
\textbf{SKD-TSTSAN (ours)} & 7          & \textbf{59.95}    & \textbf{57.31}    \\ \hline\hline
\end{tabular}
\end{minipage}
\end{table}
input, which makes it difficult to extract the motion and temporal information of MEs. In contrast, STSTNet and TSCNN use the optical flow as input and achieve better performance. Our SKD-TSTSAN achieves the best recognition performance, reaching 77.11\% UF1 and 74.77\% UAR and showing improvements of 6.68\% and 6.01\% over TSCNN. Similar improvements are observed in the 4-category emotion classification: a 3.95\% over TSCNN for UF1 and a 2.99\% increase over MMNet for UAR.

To further validate our proposed SKD-TSTSAN, we perform comparative experiments on the large-scale ME dataset CAS(ME)$^3$ using 3, 4, and 7 emotion categories, as shown in Table \ref{table4}. In the case of 3 emotion categories, SKD-TSTSAN obtains a UF1 score of 86.48\% and a UAR score of 83.44\%, which indicates an improvement of 8.16\% in UF1 and 7.38\% in UAR compared to the second-ranked MER-CLIP \citep{liu2025mer} (78.32\% UF1 and 76.06\% UAR). For the 4 emotion categories, SKD-TSTSAN outperforms MER-CLIP by 10.96\% in UF1 and 10.15\% in UAR. For all 7 emotion categories, SKD-TSTSAN achieves 59.95\% UF1 and 57.31\% UAR, surpassing MER-CLIP (49.97\% UF1 and 50.14\% UAR) by 9.98\% and 7.17\%, respectively. When tested with 3, 4, and 7 emotion categories, SKD-TSTSAN consistently exhibits substantial improvements over all comparison methods, demonstrating its strong capability on large-scale multi-class MER tasks.

\begin{figure}[t]
\centering
\includegraphics[scale=1.3]{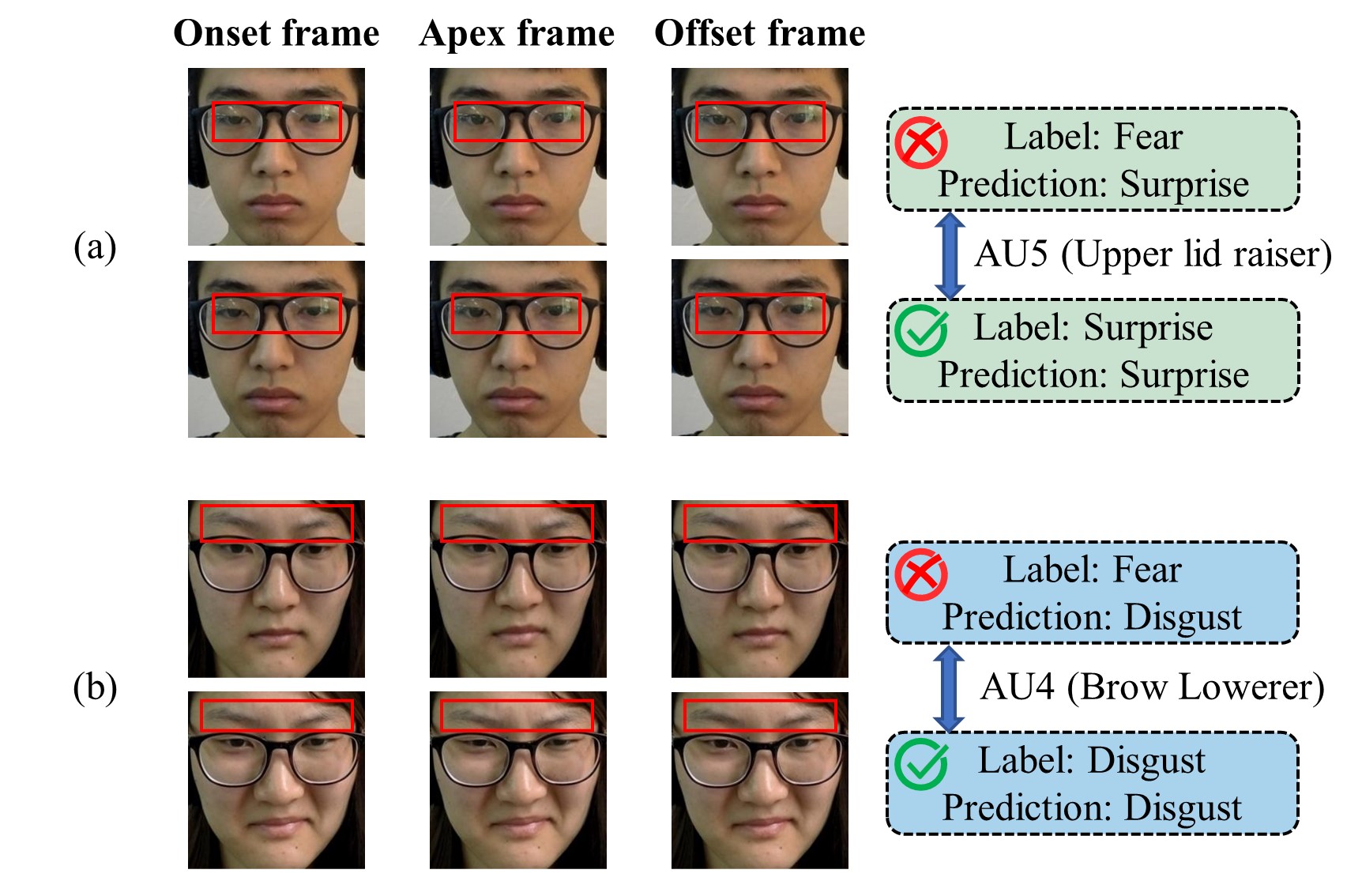}
\caption{Some correctly and incorrectly classified samples in the 6-class classification on the MMEW dataset. (a) indicates that a sample labeled as “Fear” is predicted as “Surprise”. Both it and another correctly classified sample labeled as “Surprise” appear in AU5. (b) shows a sample labeled as “Fear” is predicted as “Disgust”. Similar to another sample labeled as “Disgust”, it occurs in AU4.}\label{fig6}
\end{figure}

In addition, the confusion matrices in Fig. \ref{fig5} report the results of SKD-TSTSAN on multiple datasets. From Fig. \ref{fig5}, it can be observed that SKD-TSTSAN achieves strong performance on the 3-category MER task for the composite dataset. Specifically, the CASME II dataset shows the best recognition results, while the "Positive" category on the SAMM dataset and the "Negative" category on the SMIC dataset exhibit relatively weaker performance. On the MMEW dataset, SKD-TSTSAN performs well on the 4-class MER task. On the 6-class MER task, SKD-TSTSAN has poor recognition performance for the “Fear” and “Sadness” samples. One of the main reasons is the small number of samples for the “Fear” and “Sadness” emotion categories. Besides, different emotion categories may correspond to the same AU. For example, both “Fear” and “Surprise” expressions appear in AU5 (“Upper Lid Raiser”), while “Fear” and “Disgust” expressions both involve AU4 (“Brow Lowerer”), as shown in Fig. \ref{fig6}. Therefore, samples labeled as “Fear” are easily misclassified as “Surprise” and “Disgust” categories, which have larger sample sizes. On the CAS(ME)$^3$ dataset, the performance of SKD-TSTSAN is also affected by the imbalanced class distribution. For the 3 and 4 emotion categories, SKD-TSTSAN has poorer recognition performance on “Positive”, which has fewer samples. For the 7 emotion categories, SKD-TSTSAN performs better in the “Disgust”, “Others”, and “Surprise” categories. One of the reasons for the poor performance on the CAS(ME)$^3$ 
\begin{table}[t]
\vspace{-0.1cm} 
\footnotesize  
\setlength{\abovecaptionskip}{0.1cm}
\begin{minipage}{1.0\textwidth}
\centering
\caption{Ablation studies on the CASME II and MMEW datasets.}\label{table5}
\begin{tabular}{cccc|cc|cc}
\hline\hline
\multicolumn{4}{c|}{Methods}    & \multicolumn{2}{c|}{CASME II (5-category)} & \multicolumn{2}{c}{MMEW (6-category)} \\ \hline
Mag & ECA & TSM & SKD & UF1 (\%)             & UAR (\%)            & UF1 (\%)          & UAR (\%)          \\ \hline
              &     &     &     & 77.49                & 75.85               & 66.42             & 64.23             \\
\checkmark              &    &    &    & 87.24                & 87.64               & 67.99             & 65.08             \\
 & \checkmark & & & 80.11 & 79.83 & 67.69 & 66.42 \\
 & & \checkmark & & 83.67 & 83.01 & 66.96 & 64.46 \\
\checkmark              & \checkmark    &     &     & 88.37                & 87.76               & 70.47             & 68.97             \\
\checkmark              &     & \checkmark    &     & 88.78                & 88.85               & 69.54             & 68.90             \\
\checkmark              & \checkmark    & \checkmark    &     & {\underline{88.80} }          & {\underline {89.10}}         & {\underline {73.78}}       & {\underline {71.31}}       \\
\checkmark              & \checkmark    &  \checkmark   &  \checkmark   & \textbf{90.34}       & \textbf{89.47}      & \textbf{77.11}    & \textbf{74.77}    \\ \hline\hline
\end{tabular}
\end{minipage}
\end{table}
dataset is that the onset frame annotations of some samples are the same as the apex frame annotations. This results in no motion information in the optical flow between them, leading to misclassification.

\subsection{Analysis and Discussion}
\label{subsec4-5}
\subsubsection{Ablation Studies}
\label{subsubsec4-5-1}
In this section, to illustrate the effect of each component, we conduct ablation studies on the CASME II and MMEW datasets. Initially, a three-stream baseline network is trained using the focal loss ${{L}_{FL}}$. This baseline network is based on the TSCNN architecture but incorporates several modifications to improve performance, including the addition of batch normalization layers after each convolutional layer, replacing the last pooling operation with global average pooling, and introducing a Dropout layer before the final fully connected layer. Subsequently, we incrementally introduce submodules into the baseline model, including motion magnification (Mag) modules, efficient channel attention (ECA) modules, temporal shift modules (TSMs), and self-knowledge distillation (SKD).

As presented in Table \ref{table5}, compared to the baseline model, the introduction of motion magnification modules improves the UF1 and UAR of the 5-class MER task on the CASME II dataset by 9.75\% and 11.79\%. It also increases the UF1 and UAR of the 6-class MER task on the MMEW dataset by 1.57\% and 0.85\%, respectively. This indicates that motion magnification modules can improve MER performance by enhancing the motion intensity in each stream.

\begin{figure}[t]
\centering
\includegraphics[scale=1.2]{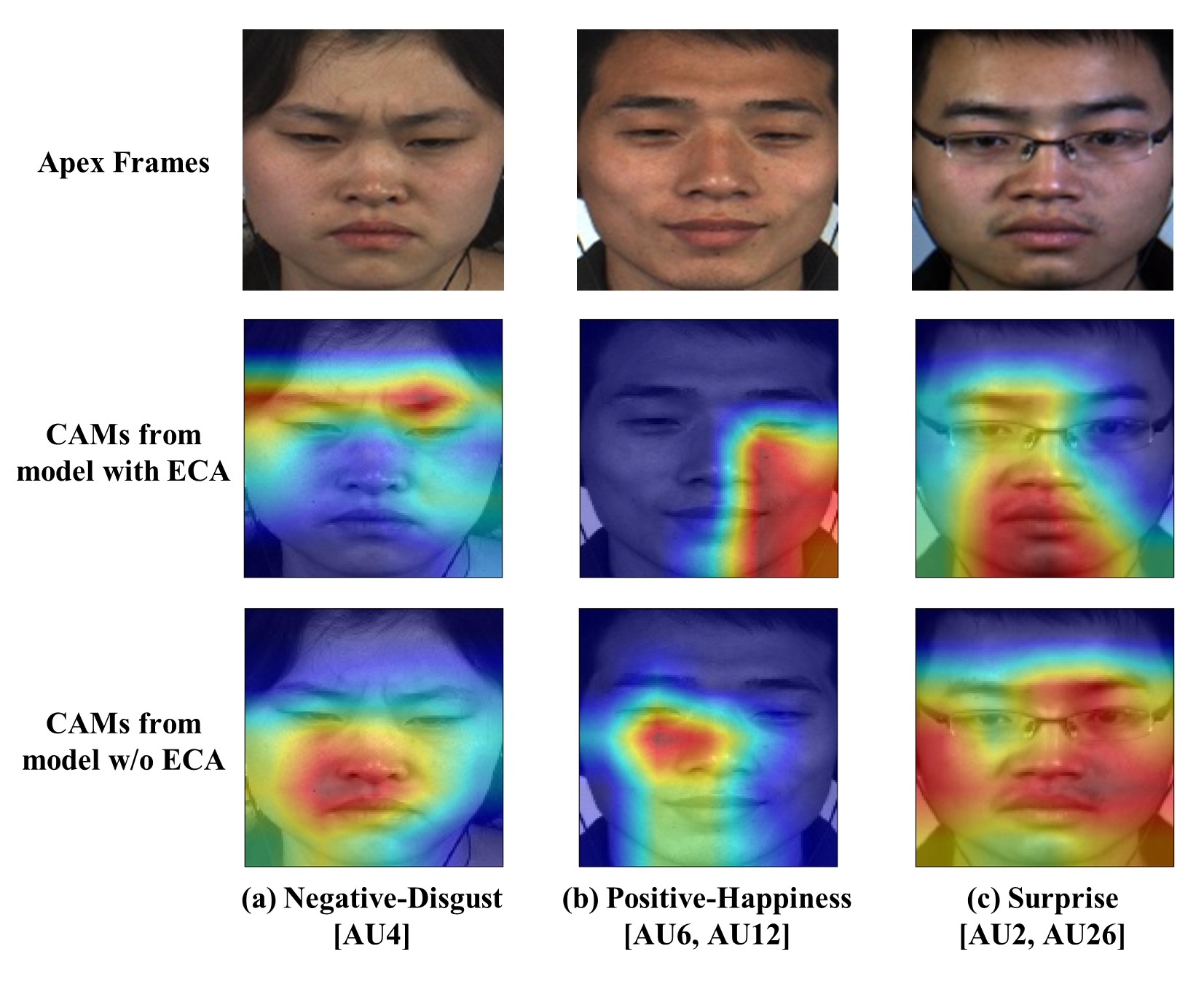}
\caption{Visualization of class activation maps (CAMs) of three ME samples from the CASME II dataset. The first to third rows represent the apex frames, and CAMs from networks with and without ECA modules, respectively. The emotion label for (a) is “Negative (Disgust)”, reflected in AU4 (“Brow Lowerer”). The emotion label for (b) is “Positive (Happiness)”, appearing in AU6 (“Cheek Raiser”) and AU12 (“Lip Corner Puller”). The emotion label for (c) is “Surprise”, reflected in AU2 (“Outer Brow Raiser”) and AU26 (“Jaw Drop”).}\label{fig7}
\end{figure}

We further introduce a channel attention mechanism by incorporating ECA modules into the L-stream of the network. Compared to the baseline, the ECA module improves UF1 and UAR by 2.62\% and 3.98\% on the CASME II dataset, and by 1.27\% and 2.19\% on the MMEW dataset. When combined with the motion magnification module, the ECA module further enhances performance, improving UF1 and UAR by 1.13\% and 0.12\% on the CASME II dataset, and 2.48\% and 3.89\% on the MMEW dataset. Furthermore, to better understand how ECA modules influence the final recognition performance, the gradient-weighted class activation mapping (Grad-CAM) \citep{selvaraju2017grad} is utilized for visualization analysis. Fig. \ref{fig7}(a), (b), and (c) represent the samples of three different emotion categories in the CASME II dataset: “Negative (Disgust)”, “Positive (Happiness)”, and “Surprise”. The apex frames are shown in the first row, and the class activation maps (CAMs) of the SKD-TSTSAN with and without the ECA module are shown in the second and third rows. These maps are from the fourth convolutional layer of the S-stream, and the red regions highlight the areas that are crucial for the MER results. According to the Facial Action Coding System (FACS), the “Disgust” expression usually involves AU9 (“Nose Wrinkler”), AU10 (“Upper Lip Raiser”), and AU4 (“Brow Lowerer”) with AU7 (“Lid Tightener”). The “Happiness” expression often involves AU6 (“Cheek Raiser”) and AU12 (“Lip Corner Puller”). The “Surprise” expression is usually found in AU5 (“Upper Lid Raiser”), AU26 (“Jaw Drop”), AU1 (“Inner Brow Raiser”), and AU2 (“Outer Brow Raiser”). As seen in the CAMs in Fig. \ref{fig7}(a), (b), and (c), when ECA modules are used, the facial regions with strong activation are consistent with the AUs defined by FACS. For instance, in Fig. \ref{fig7}(a), the eyebrow region is highlighted; in Fig. 7(b), the cheek and lip corner regions are emphasized; in Fig. \ref{fig7}(c), the eyebrow and chin regions are highlighted. However, when the ECA module is not used, the model tends to focus on regions less relevant to the expressions. For example, the model without ECA modules focuses on the cheek area in Fig. \ref{fig7}(c), which is not relevant to the “Surprise” expression. Therefore, by incorporating ECA modules into the L-stream, we not only suppress redundant information in the L-stream input but also enable the S-stream to focus
more on the key regions of ME muscle movements. Although ECA operates within the L-stream and does not share parameters with the S-stream, the three streams are jointly optimized through a shared classification objective. When ECA suppresses task-irrelevant patches in the L-stream and amplifies expression-relevant ones, the shared loss function propagates gradients that guide all streams—including the S-stream—to align their spatial attention with the same discriminative facial regions. This cross-stream regularization effect is reflected in the S-stream's CAMs: the model with ECA consistently highlights regions that correspond to the expression-specific AUs, whereas the model without ECA shows diffuse and less structured activation.

\begin{figure}[t]
\centering
\includegraphics[scale=1.2]{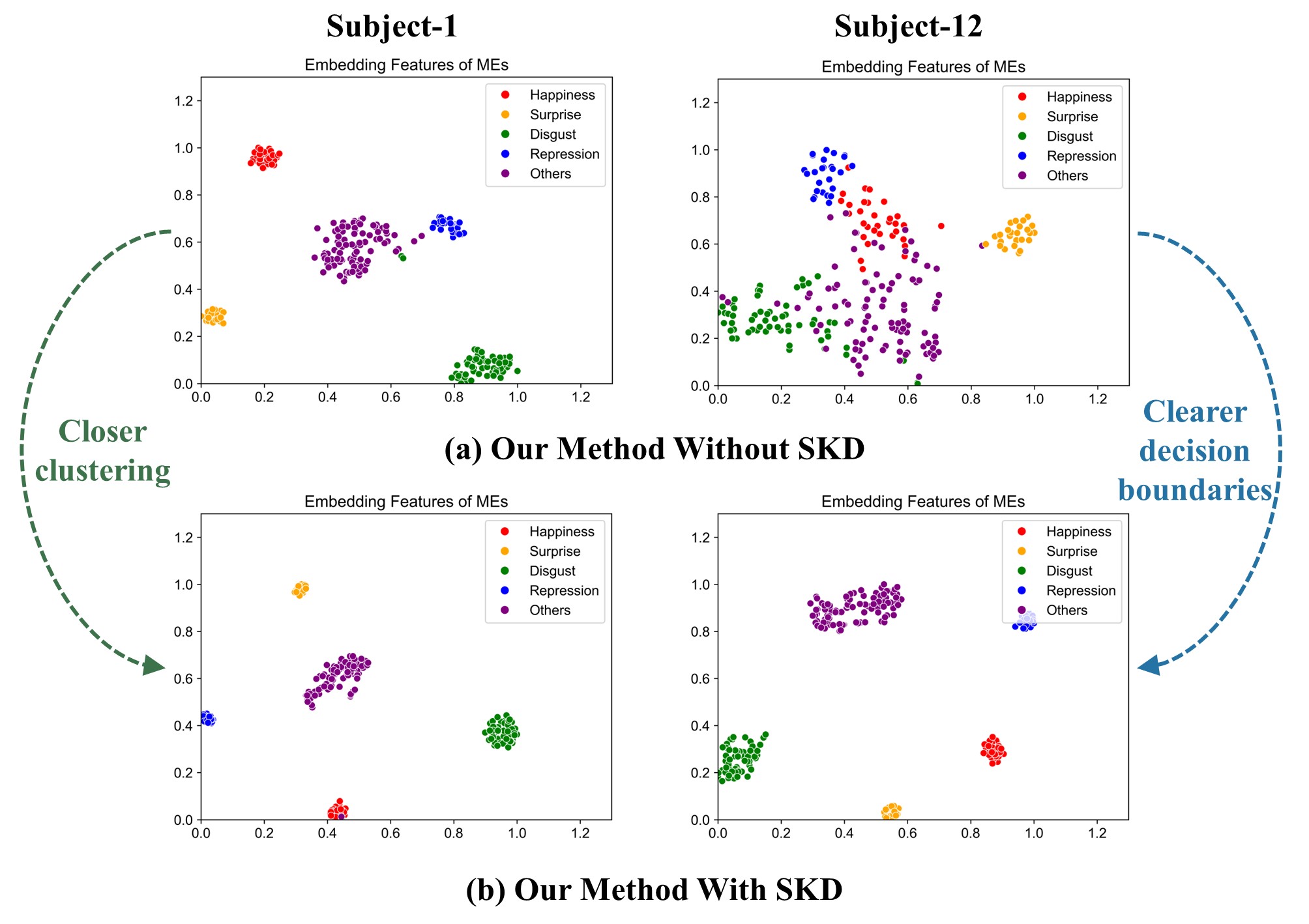}
\caption{The t-SNE visualization of ME features generated by: (a) Our method without SKD; (b) SKD-TSTSAN.}\label{fig8}
\end{figure}

Next, the introduction of TSM in the T-stream enables the exchange of temporal information between adjacent frames. Compared to the baseline, the TSM improves UF1 and UAR by 6.18\% and 7.16\% on the CASME II dataset, and by 0.54\% and 0.23\% on the MMEW dataset. These results demonstrate the effectiveness of TSM in capturing temporal dynamics. When combined with the motion magnification module, the TSM further enhances performance on CASME II and MMEW datasets. Additionally, when both ECA modules and TSMs are used, the performance is better than using only one of them, demonstrating the complementarity of the introduced modules. Furthermore, when the self-knowledge distillation (SKD) is introduced into MER, UF1 and UAR on CASME II are improved by 1.54\% and 0.37\%, and UF1 and UAR on MMEW are increased by 3.33\% and 3.46\%, respectively. For an intuitive analysis of the effect of SKD, as shown in Fig. \ref{fig8}, t-distributed stochastic neighbor embedding (t-SNE) is employed to visualize the learned ME features of subjects 1 and 12 in the CASME II dataset. Fig. \ref{fig8}(a) and (b) illustrate the embedding ME features of the network without and with SKD, respectively. It can be easily observed that with the addition of SKD, the clustering within the same category is closer, and the decision boundaries between different categories are clearer, thus the network has better
performance. In addition, the deepest section of the network serves as the teacher model in SKD without the need to design and train an additional teacher model, which reduces the training time.

\subsubsection{More Analysis on the Effect of Transfer Learning}
\label{subsubsec4-5-2}
In Fig. \ref{fig9}(a), we provide the MER results obtained by training our model on the CASME II and SAMM datasets without and with transfer learning (TL) from the macro-expression dataset CK+. For the CASME II dataset, the introduction of transfer learning improved the UAR of the 5-class MER task by 1.86\%. For the SAMM dataset, transfer learning enhanced the UAR of the 3-class MER task by 8.89\%. This improvement is attributed to the fact that macro-expressions and micro-expressions share similar AUs in conveying emotions. Consequently, we employ transfer learning in all comparative experiments on ME datasets.

\subsubsection{More Analysis on the Choice of Amplification Factors}
\label{subsubsec4-5-3}
\begin{figure}[t]
\centering
\includegraphics[scale=0.95]{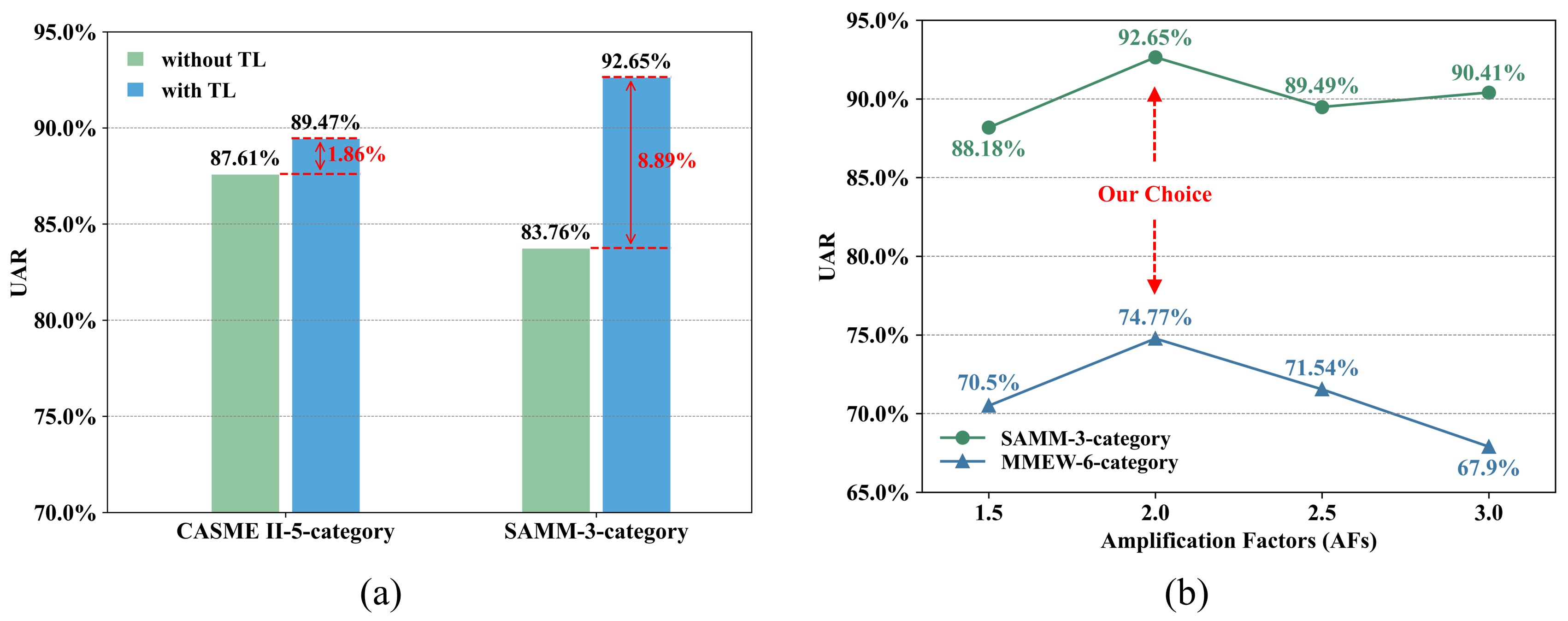}
\caption{UAR performance of SKD-TSTSAN under different experimental configurations. (a) Comparison results with and without transfer learning on the CASME II and SAMM datasets. (b) Comparison results with different amplification factors on the SAMM and MMEW datasets.}\label{fig9}
\end{figure}

To evaluate the effect of different amplification factors (AFs) on the performance, we provide the UAR results obtained by training our model under different AF configurations on the SAMM and MMEW datasets in Fig. \ref{fig9}(b). For the 3-class MER task on the SAMM dataset, our SKD-TSTSAN consistently outperforms the others MER methods ($\le $81.24\% UAR) regardless of the AF value, indicating the 
robustness and effectiveness of our approach. When the AF is too small, the facial muscle movements are not obvious; while when the AF is too large, the muscle movements are excessively deformed, both of which affect the MER performance. When the AF is set to 2, the results are optimal on both the SAMM and MMEW datasets, so this value is used as our final configuration.

\subsubsection{More Analysis on the L-stream Grid Partition Granularity}
\label{subsubsec4-5-partition}
To validate the choice of $n{=}4$ in the $4\times4$ grid partition used in the L-stream, we compare three partition granularities—$2\times2$, $4\times4$, and $8\times8$—on the CASME II 5-class MER task. As shown in Table~\ref{table_partition}, the $4\times4$ grid achieves the best performance with 90.34\% UF1 and 89.47\% UAR. A coarser $2\times2$ partition yields the lowest performance, as each of the 4 large patches covers a wide facial region and thus retains significant redundant information, reducing the discriminative power of local features. An $8\times8$ partition provides more spatial detail but performs worse than $4\times4$: at this finer granularity, each small patch covers only a limited area, and subtle ME muscle movements may span multiple patch boundaries, leading to fragmented spatial features and reduced robustness. The $4\times4$ partition provides a balanced trade-off between local discriminability and spatial coherence, and is therefore adopted as our default configuration.

\begin{table}[t]
\footnotesize
\setlength{\abovecaptionskip}{0.1cm}
\begin{minipage}{1.0\textwidth}
\centering
\caption{Ablation study on L-stream grid partition granularity on the CASME II dataset with five emotion categories.}\label{table_partition}
\begin{tabular}{c|cc}
\hline\hline
\multirow{2}{*}{Grid Partition} & \multicolumn{2}{c}{Metrics} \\ \cline{2-3}
                                & UF1 (\%)     & UAR (\%)     \\ \hline
$2\times2$ (4 patches)          & 83.03        & 81.72        \\
$4\times4$ (16 patches, ours)   & \textbf{90.34}        & \textbf{89.47}        \\
$8\times8$ (64 patches)         & \underline{86.82}        & \underline{87.61}        \\ \hline\hline
\end{tabular}
\end{minipage}
\end{table}

\begin{figure}[t]
\centering
\includegraphics[scale=1.4]{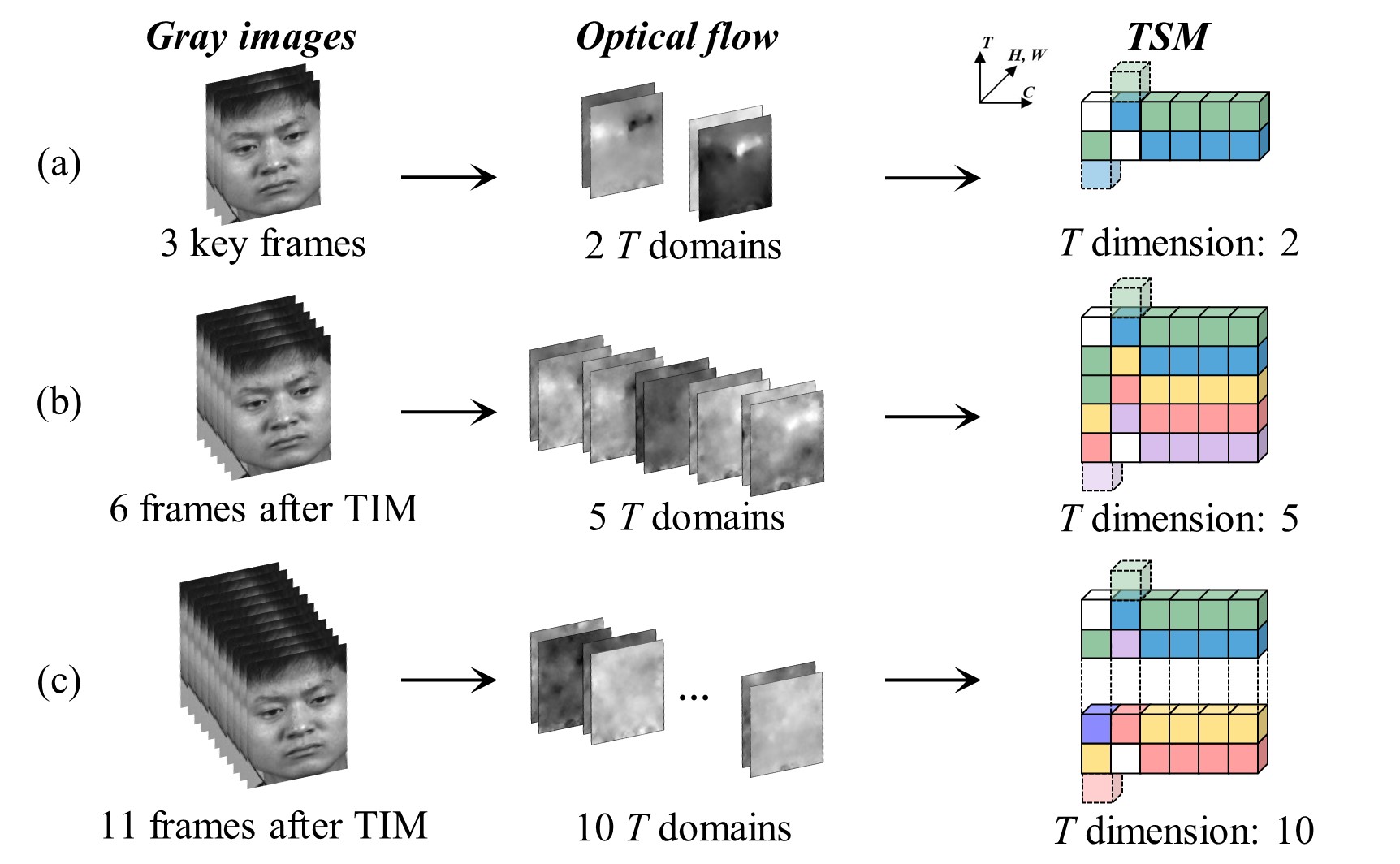}
\caption{Visualization of T-flow inputs with different temporal domain distributions.}\label{fig10}
\end{figure}
\begin{table}[t]
\footnotesize  
\setlength{\abovecaptionskip}{0.1cm}
\begin{minipage}{1.0\textwidth}
\centering
\caption{Studies on T-stream input with different temporal domain distributions on the CASME II dataset with five emotion categories.}\label{table6}
\begin{tabular}{c|cc}
\hline\hline
\multirow{2}{*}{Number of Temporal domains} & \multicolumn{2}{c}{Metrics} \\ \cline{2-3} 
                                            & UF1 (\%)     & UAR (\%)     \\ \hline
2                                           & \textbf{90.34}        & \textbf{89.47}        \\
5                                           & 86.10         & 86.22        \\
10                                          & \underline{87.38}        & \underline{86.67}        \\ \hline\hline
\end{tabular}
\end{minipage}
\end{table}

\subsubsection{More Analysis on the Temporal Domain Distribution}
\label{subsubsec4-5-4}
As illustrated in Fig. \ref{fig10}(a), in our comparative experiment configuration, our T-stream inputs are optical flow maps from two different temporal domains, calculated from the onset frame, apex frame, and offset frame. To explore the effect of different temporal distributions of T-stream inputs, as shown in Fig. \ref{fig10}(b) and (c), we first interpolate each ME sequence to 6 and 11 frames using the temporal interpolation model (TIM) \citep{pfister2011recognising}. The optical flow maps are then calculated between adjacent frames, representing the motion information from 5 and 10 different temporal domains, respectively. Subsequently, the optical flow maps from different temporal domains are input into the T-stream, where temporal modeling is performed using TSM. Table \ref{table6} provides the performance of our SKD-TSTSAN on the CASME II dataset under different temporal domain distributions. It's clear that the optimal results are obtained when we use the optical flow between the three key-frames as T-stream input, with an improvement of 2.96\% for UF1 and 2.80\% for UAR compared to inputting 10 optical flow maps. This result is consistent with the principled design rationale behind our two-domain construction. The onset, apex, and offset frames are the three semantically anchored key frames of a ME, marking the start, peak, and end of the expression. The onset-to-apex flow captures the muscle activation phase, while the apex-to-offset flow captures the deactivation phase; together, they encode the complete motion trajectory of the ME with maximum per-domain signal amplitude, since each flow map spans the full range of motion within its respective phase. In contrast, denser temporal sampling via TIM distributes this motion across more intervals, reducing the per-interval motion magnitude and producing less discriminative, noisier flow maps.

\subsubsection{More Analysis on the Temporal Shift Direction}
\label{subsubsec4-5-tsm}
To investigate the contribution of temporal shift direction and to evaluate the practicality of bidirectional temporal modeling, we compare three variants of the T-stream on the CASME II 5-class MER task: (1) \textit{No TSM}, where the temporal shift module is removed entirely; (2) \textit{Uni-directional TSM}, where only past-to-current information flow is retained (only the forward shift is applied); and (3) \textit{Bi-directional TSM} (our default), where both forward and backward channel shifts are applied.

As shown in Table~\ref{table_tsm}, removing TSM entirely yields the lowest performance (85.83\% UF1, 84.90\% UAR), confirming the necessity of temporal feature exchange for MER. The uni-directional variant achieves 88.92\% UF1 and 89.34\% UAR, substantially outperforming the no-TSM baseline and demonstrating that even one-way temporal information propagation is beneficial. The bidirectional TSM further improves UF1 to 90.34\%, showing that mutual information exchange between the two temporal domains provides complementary cues that uni-directional modeling cannot capture.

We note that in fully online or real-time scenarios, future frames (such as the offset stage) may not yet be available, which would additionally require causal input construction rather than solely a unidirectional shift. The uni-directional TSM result here provides a useful reference: it shows that the temporal modeling component alone brings substantial gains, while the bidirectional design is justified for the offline processing setting of this work.

\begin{table}[t]
\footnotesize
\setlength{\abovecaptionskip}{0.1cm}
\begin{minipage}{1.0\textwidth}
\centering
\caption{Ablation study on temporal shift direction on the CASME II dataset with five emotion categories.}\label{table_tsm}
\begin{tabular}{c|cc}
\hline\hline
\multirow{2}{*}{TSM Variant} & \multicolumn{2}{c}{Metrics} \\ \cline{2-3}
                              & UF1 (\%)     & UAR (\%)     \\ \hline
No TSM                        & 85.83        & 84.90        \\
Uni-directional TSM           & \underline{88.92}        & \underline{89.34}        \\
Bi-directional TSM (ours)     & \textbf{90.34}        & \textbf{89.47}        \\ \hline\hline
\end{tabular}
\end{minipage}
\end{table}

\subsubsection{More Analysis on the Performance of Different Classifiers}
\label{subsubsec4-5-5}
\begin{table}[t]
\scriptsize  
\setlength{\abovecaptionskip}{0.1cm}
\centering
\caption{Experimental results of different classifiers on multiple datasets.}\label{table7}
\begin{tabular}{c|c|cccccc}
\hline\hline
\multirow{2}{*}{Dataset}  & \multirow{2}{*}{\# Classes} & \multicolumn{2}{c}{Classifier 1/3} & \multicolumn{2}{c}{Classifier 2/3} & \multicolumn{2}{c}{Classifier 3/3} \\ \cline{3-8} 
                          &                             & UF1 (\%)         & UAR (\%)        & UF1 (\%)         & UAR (\%)        & UF1 (\%)         & UAR (\%)        \\ \hline
Full & \multirow{4}{*}{3} & 76.61 & 78.49 & 85.95 & 86.74 & 90.20 & 90.83 \\
CASME II &   & 81.06 & 82.66 & 90.16 & 89.86 & 98.58 & 98.58 \\
SAMM &   & 71.42 & 70.82 & 83.24 & 81.63 & 87.70 & 84.00 \\
SMIC &   & 73.44 & 75.54 & 82.81 & 84.22 & 84.16 & 85.71 \\ \hline
CASME II & 5                           & 75.00               & 74.61           & 87.80             & 87.98           & 90.34            & 89.47           \\ \hline
\multirow{2}{*}{MMEW}     & 6                           & 56.21            & 54.56           & 72.64            & 69.81           & 77.11            & 74.77           \\
                          & 4                           & 71.43            & 71.96           & 88.47            & 88.33           & 90.41            & 90.44           \\ \hline
\multirow{3}{*}{CAS(ME)$^3$} & 3                           & 68.97            & 66.35           & 83.19            & 80.99           & 86.48            & 83.44           \\
                          & 4                           & 52.39            & 51.68           & 73.27            & 70.14           & 76.40             & 72.57           \\
                          & 7                           & 39.09            & 38.86           & 54.42            & 52.48           & 59.95            & 57.31           \\ \hline\hline
\end{tabular}
\end{table}

\begin{figure}[t]
\centering
\includegraphics[scale=1.2]{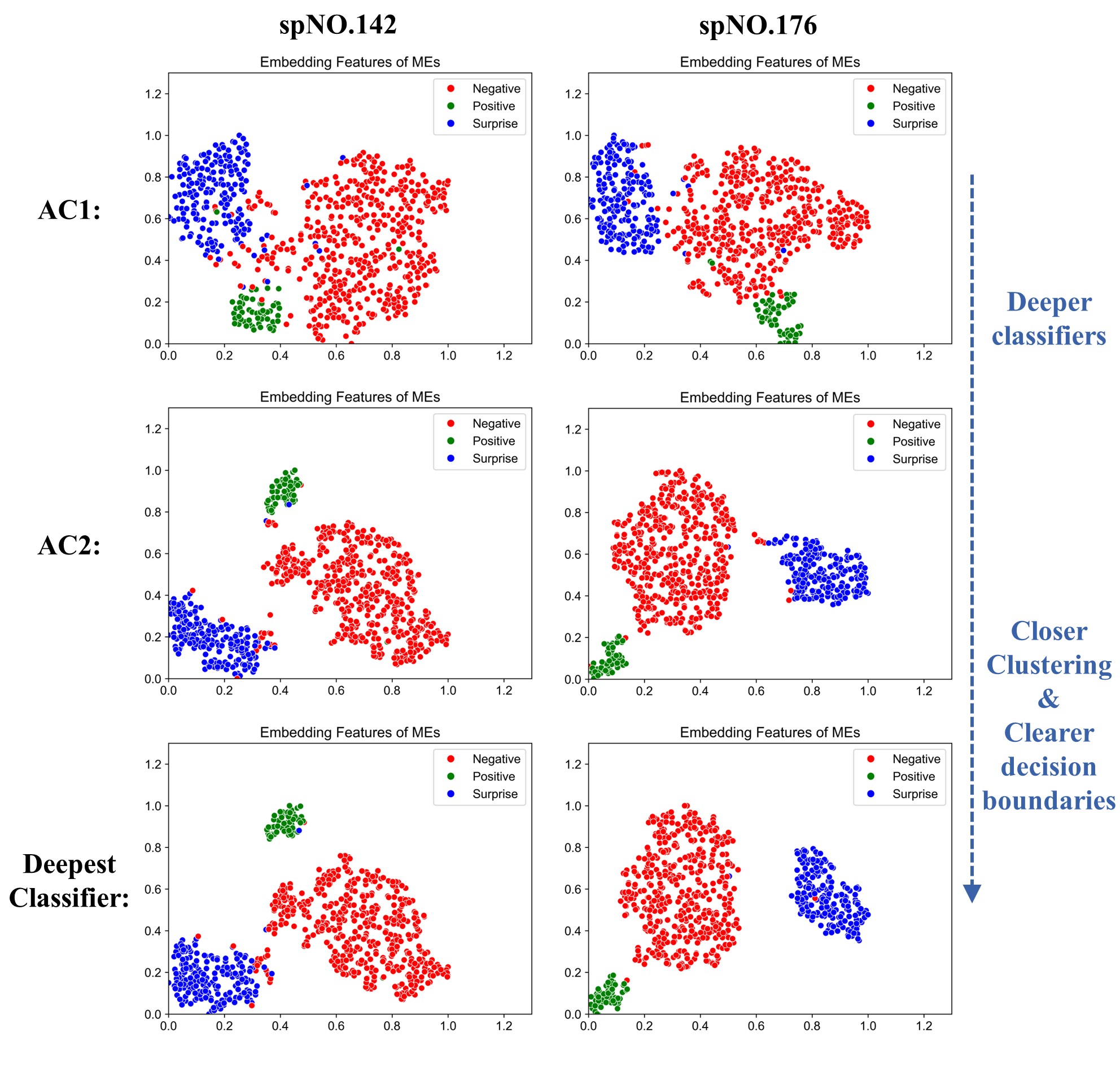}
\caption{The t-SNE results of ME features generated by AC1, AC2, and the deepest classifier. The first and second columns are derived from subjects spNO.142 and spNO.176 in the CAS(ME)$^3$ dataset, respectively.}\label{fig11}
\end{figure}

To verify that deeper classifiers in SKD extract more discriminative features, we compare the performance of each classifier on multiple datasets. As shown in Table \ref{table7}, the deepest classifier 3/3 achieves the highest UF1 and UAR scores on different datasets. The deeper classifier 2/3 performs better than the shallow classifier 1/3, since the features extracted from the deeper section are more discriminative. Although classifier 2/3 does not perform as well as the deepest classifier 3/3, its performance on different datasets is comparable to or even better than that of other advanced MER methods, indicating the effectiveness of our SKD-TSTSAN. To visualize the discriminative features extracted by different classifiers, t-SNE is utilized to visualize the embedding ME features of subjects spNO.142 and spNO.176 in the CAS(ME)$^3$ dataset, as shown in Fig. \ref{fig11}. It is obvious that the deeper classifiers have more concentrated clustering of the same category and clearer decision boundaries between different categories, indicating that more discriminative features are extracted.

\subsubsection{Computational Complexity Analysis}
\label{subsubsec4-5-complexity}
To assess the computational efficiency of SKD-TSTSAN, Table~\ref{table_complexity} reports parameters (Params) and FLOPs for several representative methods. We distinguish between SKD-TSTSAN Full (training mode, with two auxiliary classifiers) and SKD-TSTSAN Inference (deployment mode, ACs removed).

At inference, SKD-TSTSAN requires only 1.611M parameters—substantially fewer than the TSCNN baseline (10.760M), MMNet (10.075M), and HTNet (114.431M)—owing to the use of global average pooling in place of large fully connected layers. Although the FLOPs are higher than the TSCNN baseline due to the motion magnification and ECA/TSM components, they remain well below MMNet (14.345G). STSTNet is extremely lightweight but achieves much lower recognition accuracy. These results confirm that SKD-TSTSAN achieves a favorable balance between recognition performance and model complexity, and the SKD auxiliary classifiers add no overhead at inference.

\begin{table}[t]
\footnotesize
\setlength{\abovecaptionskip}{0.1cm}
\centering
\caption{Computational complexity comparison.}\label{table_complexity}
\begin{tabular}{lcc}
\hline\hline
Model & Params & FLOPs \\ \hline
STSTNet \citep{liong2019shallow}                & 2.485 K   & 897.6 K  \\
TSCNN Baseline \citep{song2019recognizing}        & 10.760 M  & 458.4 M  \\
MMNet \citep{li2022mmnet}                       & 10.075 M  & 14.345 G \\
HTNet \citep{wang2024htnet}                     & 114.431 M & 378.310 M \\
SKD-TSTSAN Full (ours)                          & 2.946 M   & 1.816 G  \\
SKD-TSTSAN Inference (ours)                     & \textbf{1.611 M}   & 1.704 G  \\ \hline\hline
\end{tabular}
\end{table}

\subsubsection{Cross-Dataset Generalization}
\label{subsubsec4-5-cross}
To evaluate cross-dataset generalization, we train on a subject-split partition of CAS(ME)$^3$ and test on the complete SMIC dataset using the shared 3-class label space (Positive, Negative, Surprise), without any fine-tuning on SMIC. As shown in Table~\ref{table_cross}, SKD-TSTSAN achieves 51.24\% UF1 and 56.00\% UAR, improving over the TSCNN baseline by 6.86\% UF1 and 6.99\% UAR. This demonstrates that the motion magnification modules and self-knowledge distillation help the model learn more transferable ME representations that generalize across datasets with different recording conditions (frame rate, resolution, and subject demographics).

\begin{table}[t]
\footnotesize
\setlength{\abovecaptionskip}{0.1cm}
\centering
\caption{Cross-dataset generalization: trained on CAS(ME)$^3$, tested on SMIC (3-class).}\label{table_cross}
\begin{tabular}{lcc}
\hline\hline
Method & UF1 (\%) & UAR (\%) \\ \hline
TSCNN Baseline \citep{song2019recognizing} & 44.38 & 49.01 \\
SKD-TSTSAN (ours)                          & \textbf{51.24} & \textbf{56.00} \\ \hline\hline
\end{tabular}
\end{table}

\subsubsection{Evaluation on the DFME Dataset}
\label{subsubsec4-5-dfme}
To further assess the recognition capability of SKD-TSTSAN on a large-scale ME benchmark, we conduct experiments on the DFME dataset~\citep{zhao2023dfme}, which contains 7 fine-grained emotion categories from 259 subjects collected at high frame rates of 200--500~fps. The TSCNN baseline and STSTNet are reproduced by the authors under the same preprocessing pipeline and official DFME training/testing protocol as SKD-TSTSAN. Specifically, all three methods are trained on the full DFME training set (1,856 samples) and evaluated on the official Test A (474 samples) and Test B (299 samples) splits. As shown in Table~\ref{table_dfme}, SKD-TSTSAN outperforms both reproduced baselines on all metrics across both test sets. On Test A, SKD-TSTSAN achieves 43.46\% Accuracy, 33.80\% UF1, and 34.26\% UAR, surpassing STSTNet by 4.57\% UF1 and TSCNN by 11.41\% UF1. On Test B, SKD-TSTSAN achieves 35.45\% Accuracy, 31.48\% UF1, and 33.61\% UAR, outperforming both baselines by at least 3.79\% UF1. These results demonstrate that the motion magnification modules and SKD training strategy remain effective on a large-scale benchmark with fine-grained 7-class emotion categories.

\begin{table}[t]
\footnotesize
\setlength{\abovecaptionskip}{0.1cm}
\centering
\caption{Performance comparison on the DFME dataset (7-class). The TSCNN baseline and STSTNet results are reproduced by the authors under the same preprocessing protocol and official Test A/Test B splits as SKD-TSTSAN.}\label{table_dfme}
\resizebox{\textwidth}{!}{%
\begin{tabular}{lcccccc}
\hline\hline
\multirow{2}{*}{Method} & \multicolumn{3}{c}{Test A} & \multicolumn{3}{c}{Test B} \\
\cline{2-7}
 & Acc (\%) & UF1 (\%) & UAR (\%) & Acc (\%) & UF1 (\%) & UAR (\%) \\ \hline
TSCNN Baseline \citep{song2019recognizing} & 39.24 & 22.39 & 27.48 & 31.44 & 27.69 & 29.09 \\
STSTNet \citep{liong2019shallow}           & 36.29 & 29.23 & 30.92 & 26.42 & 25.83 & 25.45 \\
SKD-TSTSAN (ours)                          & \textbf{43.46} & \textbf{33.80} & \textbf{34.26} & \textbf{35.45} & \textbf{31.48} & \textbf{33.61} \\ \hline\hline
\end{tabular}%
}
\end{table}

\section{Conclusion}
\label{sec5}
In this paper, we propose a three-stream temporal-shift attention network based on self-knowledge distillation (SKD-TSTSAN) for ME recognition. First, we employ learning-based motion magnification modules to enhance muscle motion to make the ME features more discriminative. To suppress redundant information in the local-spatial stream and focus on important facial regions, we introduce a channel attention mechanism using the ECA modules. In addition, TSMs are incorporated into the dynamic-temporal stream for temporal modeling through information exchange in the temporal dimension. Furthermore, we introduce SKD into ME recognition by adding auxiliary classifiers into the network, taking the deepest part of the network as the teacher model for the shallower part and further improving the performance. Finally, extensive experiments on the CASME II, SAMM, SMIC, MMEW, and CAS(ME)$^3$ datasets show that our SKD-TSTSAN outperforms other cutting-edge MER methods. Additional evaluations on the large-scale DFME benchmark (7-class) and a cross-dataset generalization study (trained on CAS(ME)$^3$, tested on SMIC) further validate the effectiveness and transferability of the proposed approach.

In this paper, we classify MEs into a set of emotion categories using the onset frames, apex frames, and offset frames labeled in the dataset. However, to achieve an end-to-end ME analysis system, we still need to design an ME spotting method to identify key-frames \citep{wang2025keyframe}. Therefore, in future work, we will explore ME spotting methods in long videos \citep{wang2021mesnet}. Additionally, many current research efforts treat the ME spotting and recognition tasks separately \citep{song2019recognizing}. However, these two tasks are strongly related and both require extracting subtle facial muscle motion features. Therefore, we will attempt to integrate ME spotting with our designed ME recognition network to implement ME analysis in a unified network.

\section*{Acknowledgements}
\label{sec6}
This work was funded by the Fundamental Research Funds for the Central Universities of China from the University of Electronic Science and Technology of China under Grant ZYGX2025YGLH009 and Grant ZYGX2024XJ026.












\bibliographystyle{elsarticle-num}
\bibliography{reference}

\end{document}